\documentclass[10pt,twocolumn,letterpaper]{article}

\usepackage[pagenumbers]{cvpr} 

\usepackage{graphicx}
\usepackage{amsmath}
\usepackage{amssymb}
\usepackage{booktabs}
\usepackage{caption}
\newtheorem{theorem}{Theorem}
\usepackage[misc]{ifsym}
\usepackage[accsupp]{axessibility} 
\usepackage[pagebackref,breaklinks,colorlinks]{hyperref}
\usepackage[capitalize]{cleveref}
\newcommand\blfootnote[1]{%
  \begingroup
  \renewcommand\thefootnote{}\footnote{#1}%
  \addtocounter{footnote}{-1}%
  \endgroup
}

\crefname{section}{Sec.}{Secs.}
\Crefname{section}{Section}{Sections}
\Crefname{table}{Table}{Tables}
\crefname{table}{Tab.}{Tabs.}

\def\cvprPaperID{3139} 
\def\confName{CVPR}
\def\confYear{2022}

\begin{document}
\title{Task-specific Inconsistency Alignment for Domain Adaptive Object Detection}

\author{
  Liang Zhao \quad \quad Limin Wang\textsuperscript{~\Letter}\\
  State Key Laboratory for Novel Software Technology, Nanjing University, China\\
  \small \texttt{liangzhao@smail.nju.edu.cn}, \texttt{lmwang@nju.edu.cn} \\
}
\maketitle

\begin{abstract}
Detectors trained with massive labeled data often exhibit dramatic performance degradation in some particular scenarios with data distribution gap. To alleviate this problem of domain shift, conventional wisdom typically concentrates solely on reducing the discrepancy between the source and target domains via attached domain classifiers, yet ignoring the difficulty of such transferable features in coping with both classification and localization subtasks in object detection. To address this issue, in this paper, we propose \textbf{Task-specific Inconsistency Alignment} (TIA), by developing a new alignment mechanism in separate task spaces, improving the performance of the detector on both subtasks. Specifically, we add a set of auxiliary predictors for both classification and localization branches, and exploit their behavioral inconsistencies as finer-grained domain-specific measures. Then, we devise task-specific losses to align such cross-domain disagreement of both subtasks. By optimizing them individually, we are able to well approximate the \textbf{category-} and \textbf{boundary-wise} discrepancies in each task space, and therefore narrow them in a decoupled manner. TIA demonstrates superior results on various scenarios to the previous state-of-the-art methods. It is also observed that both the classification and localization capabilities of the detector are sufficiently strengthened, further demonstrating the effectiveness of our TIA method. Code and trained models are publicly available at \url{https://github.com/MCG-NJU/TIA}.
\end{abstract}
\blfootnote{\Letter: Corresponding author.}

\section{Introduction}
\label{sec:intro}

Object detection~\cite{rcnn,fasterrcnn,maskrcnn,retina} is manifesting a high demand for massive annotated data, 
which however, due to either economic or technical reasons, is struggling to be fulfilled in some scenarios.
An alternative is to transfer knowledge from a \textit{source} domain depicting general or synthetic scenes 
to the \textit{target} domain describing particular scenes of interest.
Yet, as a consequence of the \textit{domain shift}~\cite{shift}, the performance of the detector would typically suffer dramatic degradation.
A practical strategy to cope with this dilemma is to adopt Unsupervised Domain Adaptation (UDA).
Generally, by narrowing the divergence in pixel or feature-level between the source and target domains, 
a detector trained on source labeled domain can be then well-generalized to unlabeled target domain.
This classic strategy of domain alignment, which originated from cross-domain classification~\cite{dann,adda,coda,mcd,swd,symv2}, establishes a solid foundation for downstream domain adaptive detection~\cite{daf,swda,scl,maf,htcn,umt}.

\begin{figure}
\centering
\begin{subfigure}{.32\linewidth}
  \centering
  \includegraphics[width=\linewidth]{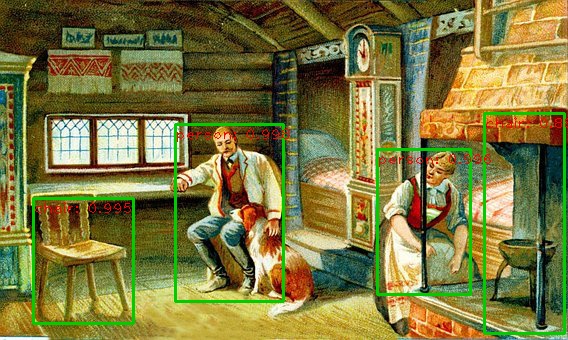}
\end{subfigure}
\begin{subfigure}{.32\linewidth}
  \centering
  \includegraphics[width=\linewidth]{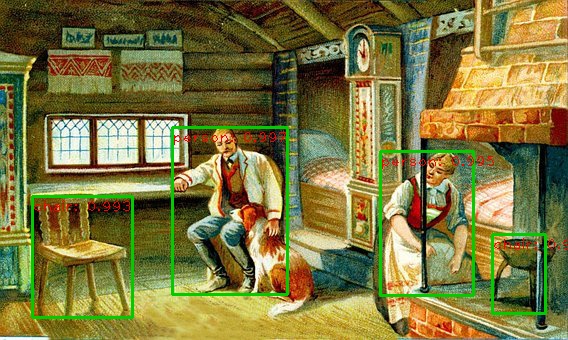}
\end{subfigure}
\begin{subfigure}{.32\linewidth}
  \centering
  \includegraphics[width=\linewidth]{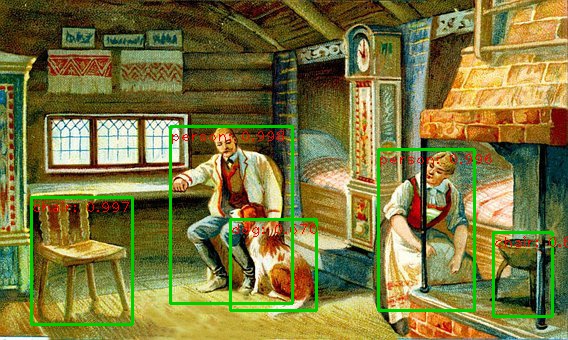}
\end{subfigure}
\\[\smallskipamount]

\begin{subfigure}{.32\linewidth}
  \centering
  \includegraphics[width=\linewidth]{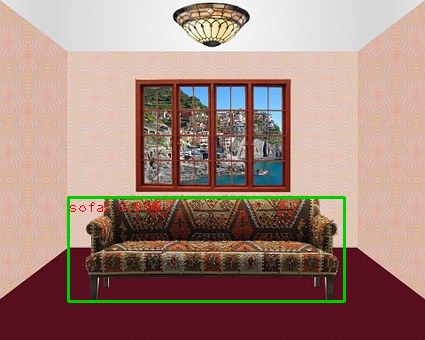}
  \caption{Vanilla~\cite{fasterrcnn}}
\end{subfigure}
\begin{subfigure}{.32\linewidth}
  \centering
  \includegraphics[width=\linewidth]{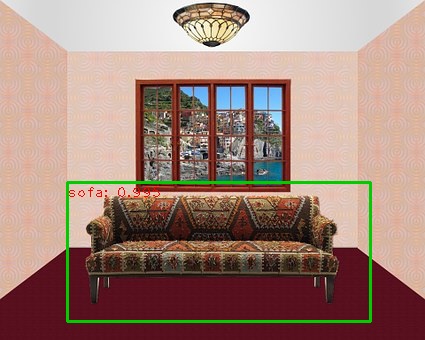}
  \caption{UMT~\cite{umt}}
\end{subfigure}
\begin{subfigure}{.32\linewidth}
  \centering
  \includegraphics[width=\linewidth]{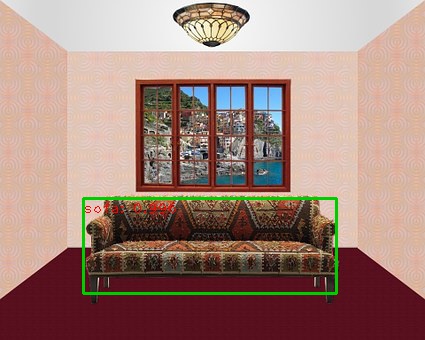}
  \caption{Our TIA}
\end{subfigure}
\\[\smallskipamount]
\caption{
Exampled images from PASCAL VOC~\cite{pascal} $\rightarrow$ Clipart~\cite{dt&pl}.
Compared with vanilla detector~\cite{fasterrcnn}, both UMT~\cite{umt} and TIA identify more foreground objects (Row 1),
yet deliver \textit{lower} as well as \textit{higher} quality bounding boxes (Row 2), respectively.
}
\label{fig1}
\vspace*{-0.4cm}
\end{figure}

Often, as an extension of domain adaptive classifiers, existing domain adaptive detectors
focus solely on decreasing the generalization error of their classifiers.
Yet, they tend to ignore the \textbf{potential improvement} of their localization errors~\cite{daf,regress}.
As shown in~\cref{fig1}, compared to vanilla detector,
it is observed that the state-of-the-art domain adaptive detector
(\ie UMT~\cite{umt}) 
is capable of correctly identifying and classifying more foreground objects,
but delivering relatively lower quality bounding boxes for them.
One possible reason is that, 
by applying domain alignment via an external binary classifier, 
the resulted transferable (\ie cross-domain invariant) features grown in the classification space
might be harmful for the localization in regression space.
Intuitively, the regression space is usually continuous and sparse and has no obvious decision boundaries, hence significantly differs from the classification space.

Motivated by this observation, we argue that 
\textit{the transferable features induced by previous adaptive detectors fail to cope well with 
both classification and localization subtasks}.
Therefore this paper for the first time, explicitly develops the feature alignment in separate task spaces,
in order to seek consistent performance gains on both classification and localization branches.
Prevalent two-stage detectors generate a single coupled region of interest (ROI) feature for both subtasks, 
hindering us from directly applying conventional alignment for each task's feature separately.
To overcome this issue, we resort to build multiple auxiliary classifiers and localizers and introduce their behavioral inconsistencies to constitute two task-specific discriminators. In this way, we are able to realize a new decoupled and fine-grained feature alignment by optimizing them separately.

Specifically, we design a general \textbf{Task-specific Inconsistency Alignment} (TIA) module to exploit the inconsistency among 
these new auxiliary predictors and apply it to both subtasks of detectors.
Therein, two task-specific losses are devised so that the behavioral disagreement among predictors can be better perceived and easily optimized.
In particular, for classification, 
we establish a stable approximation to the diversity of auxiliary classifiers' decision boundaries with the aid of Shannon Entropy (SE),
for effectively shrinking the cross-domain \textbf{category-wise} discrepancies.
Meanwhile for localization, in consideration of the continuity and sparsity of the regression space,
we leverage the Standard Deviation (SD) practically to harvest the ambiguity of various localizers' predictions at each boundary.
This allows the \textbf{boundary-wise} perception of localizers to be efficiently promoted.
Overall, by maximining these two losses, we are able to directly perform inconsistency alignments independently in fully decoupled task spaces,
thereby consistently advancing the transferability of features for both classification and localization tasks.

In summary, our contributions fall into threefold: 
(1) We empirically observe that the resulted features guided by existing feature alignment methods fail to improve the performance of both classification and localization tasks in domain adaptive object detection. To the best of our knowledge, we are the first to address this dilemma by developing domain adaptation into these two branches and directly performing alignment in these two task spaces (not feature space) independently.
(2) To effectively perform alignment in task spaces, we propose to build a set of auxiliary predictors and use their behavioral inconsistency for cross-domain alignment. These new inconsistency measures are task-specific and finer-grained, thus expected to better capture the domain difference.
(3) Exhaustive experiments have been conducted on various domain shift scenarios, demonstrating superior performance of our framework over state-of-the-art domain adaptive detectors. As shown in Fig. \ref{fig1} (c), our TIA makes significant progress in both tasks.

\section{Related Work}

\textbf{Unsupervised Domain Adaptation (UDA).}
In light of the basic assumption~\cite{theory}, extensive domain adaption methods
have been proposed~\cite{dann,adda,coda,mcd,swd,symv2}, aiming at learning transferable features to shrink the discrepancy across domains.
Recently, several methods~\cite{coda,mcd,swd,symv2} have embraced 
the \textit{consensus regularization}~\cite{consensus} strategy derived from semi-supervised learning.
Generally, multiple classifiers with varying initializations are introduced and the inconsistency among their outputs are viewed as
an \textit{indicator}, for measuring the divergence between domains.
In this way, \cite{coda} reduces this disagreement and diversifies the constructed multiple feature embeddings at the same time.
\cite{mcd} then simplifies this procedure, by iteratively maximizing and minimizing the disagreement.
On top of them, \cite{swd} introduces the wasserstein metric for mining the natural notion of dissimilarity among predictions, 
while \cite{symv1,symv2} extend the form of \cite{mcd} and explore in detail the scoring disagreement in the multi-class case.
These methods are further generalized to downstream domain adaptation tasks, 
including semantic segmentation~\cite{clan,mrnet} and keypoint detection~\cite{regress1,regress}.
In contrast, object detection is a more challenging task in that it is structurally complex and requires the simultaneous optimization of two unparalleled subtasks.
Hence, our TIA delves into the \textit{task-specific alignment}
and investigates in depth how to accurately bound then reduce both the category-wise disparities and boundary-wise ambiguity within individual task spaces.

\begin{figure*}
\centering
\includegraphics[width=\linewidth]{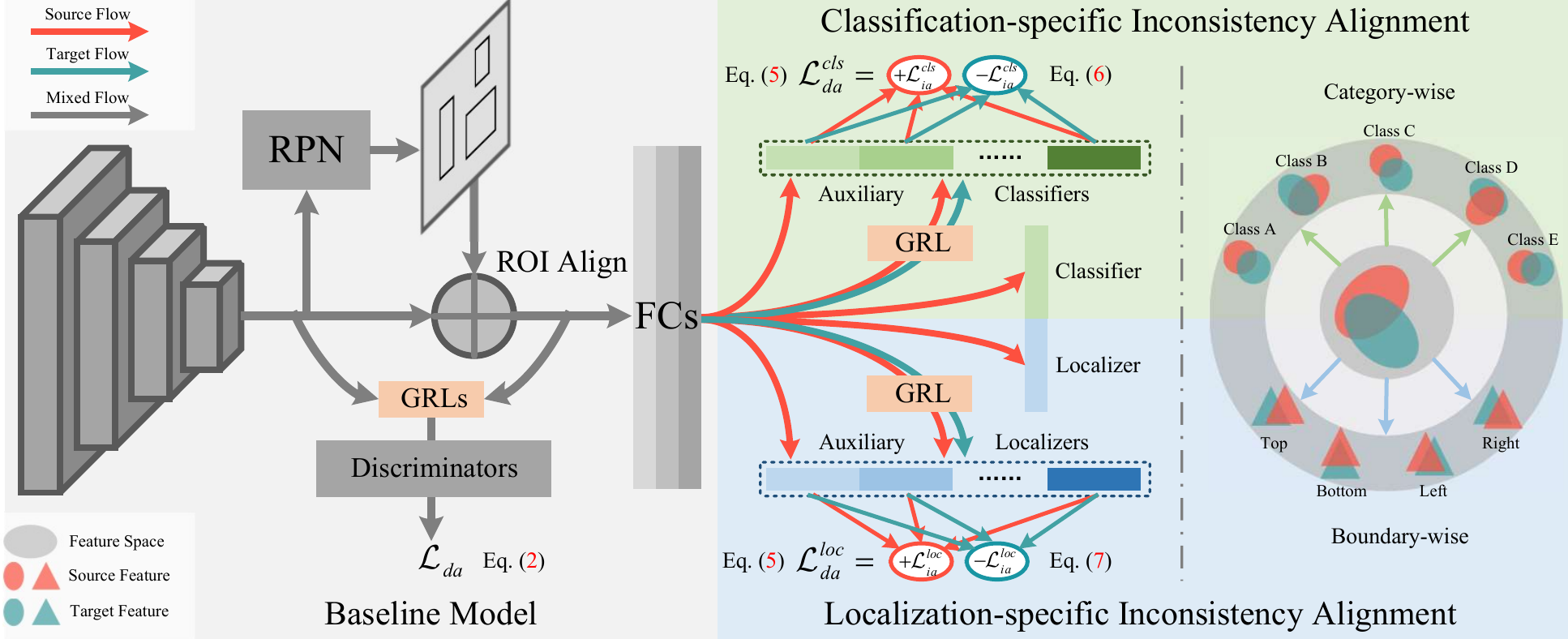}
\caption{\textbf{Framework overview.} Best viewed in color.
We develop the high-level feature alignment into separate task spaces,
by applying the proposed Task-specific Inconsistency Alignment module to 
both the classification (\textit{green} part) and localization (\textit{blue} part) branches
of the baseline detector (\textit{gray} part).
In each branch, the behavioral inconsistency of multiple auxiliary predictors is optimized
via the corresponding inconsistency-aware loss,
for essentially bridging the category-wise or boundary-wise margins between domains.
}
\label{fig2}
\end{figure*}

\textbf{UDA for Object Detection.}
Along the lines of domain adaptive classifiers, the focus of domain adaptive detectors 
is mostly on bridging the pixel or feature-level divergence between the two domains.
Many methods~\cite{dt&pl,pda,htcn,umt,rpnpa} leverage the labeled target-like images generated by CycleGAN~\cite{cyclegan} to pursue a pixel-level consistency. 
Yet far more methods~\cite{daf,swda,scl,maf,htcn,umt} are devoted to incrementally reinforcing a feature-level consistency.
Nearly all of them explicitly integrate
domain-adversarial neural network
~\cite{dann} into the detector, thereby accomplishing feature alignment
with simply domain classifiers.
\cite{daf} initially carries out domain alignment on both backbone features (image-level) and ROI features (instance-level).
After that, massive methods~\cite{swda,scl,maf,htcn,umt} continuously strengthen these two alignments,
and further improve the performance of the detector with multi-scale~\cite{maf}, contextual~\cite{swda,htcn}, spatial attention~\cite{sap}, category attention~\cite{mega}
and cross-domain topological relations~\cite{dbgl}
information.
In addition, \cite{cst} and~\cite{rpnpa} concentrate on enhancing the cross-domain performance of the region proposal network (RPN) to generate high-quality ROIs, 
whereby the former enforces collaborative training with~\cite{mcd} and self-training on RPN and region proposal classifier,
yet the latter construct one set of learnable RPN prototypes for alignment.
Problematically, almost all of existing domain adaptive detectors specialize in regulating decision boundaries of classifiers within detectors,
yet ignore the behavioral anomalies of their localizers.
In contrast, our TIA first takes this problem into account and develops the general feature alignment into independent task spaces,
leading to guaranteed accuracy for each label predictor.

\section{Methodology}

Following the regular settings of unsupervised domain adaption, 
we define a labeled source domain $\mathcal{D}_s$ and an unlabeled target domain $\mathcal{D}_t$. 
Our goal is to establish a knowledge transfer from $\mathcal{D}_s$ to $\mathcal{D}_t$ for object detection,
with a favorable generalization over the target domain being guaranteed.
In this section, we present technical details of the proposed framework,
its overall architecture is illustrated in~\cref{fig2}.
We first briefly review the baseline model (left \textit{gray} part) and, on top of it, 
thoroughly describe the proposed task-specific inconsistency alignment
(right \textit{blue} and \textit{green} parts).
In the end, some theoretical insights will be raised to explain how our method functions to improve the transferability of both subtasks within the detector.

\subsection{Baseline Model}\label{baseline}

Our framework is implemented on the basis of the popular two-stage detector Faster R-CNN~\cite{fasterrcnn}, and
the \textit{gray} areas in~\cref{fig2} represent the detector's core structure. 
Images from both domains are firstly fed into the backbone to yield image-level features, 
followed by RPN to deliver plentiful proposals,
which are then aggregated with the backbone features through ROI Align~\cite{maskrcnn} to generate a certain number of ROIs.
With the two ROI predictors on the right of FCs, the total detection loss can be formally defined as
\begin{equation}
\mathcal{L}_{det} = \mathcal{L}_{rpn} + \mathcal{L}_{roi}.
\end{equation}

To pursue the semantic consistency for subsequent modules,
we adhere to the mainstream practice of aligning features on the source and target domains,
at both mid-to-upper layers of the backbone (\ie image-level) and ROI layer (\ie instance-level).
Similar to~\cite{daf,scl,htcn}, all these feature alignments are realized by adversarial training, in terms of the domain-adversarial neural network (DANN)~\cite{dann}.
Specifically, features are conveyed via a Gradient Reversal Layer (GRL) to the discriminator $D_k$ 
for distinguishing their domain labels. The objective is as follows:
\begin{equation}\footnotesize
    \mathcal{L}_{da}= \sum_{k=1}^{K}\bigg(\frac{1}{n_s}\sum_{i=1}^{n_s}{L}\Big(D_k\big(f_{k,i}\big),d_{k,i}^s\Big)+\frac{1}{n_t}\sum_{i=1}^{n_t}{L}\Big(D_k\big(f_{k,i}\big),d_{k,i}^t\Big)\bigg),    
\end{equation}
where ${L}$ is normally a binary cross-entropy loss,
$f_{k,i}$ denotes the $i$-th feature output from the $k$-th layer and $d_{k,i}$ indicates its corresponding domain label,
$n_s$ and $n_t$ refer to the total number of features within a mini-batch in source and target domains, respectively,
and $K$ represents the total number of feature alignments.
After the above domain adaptation loss being minimized, 
the sign of gradient back-propagated from discriminator to generator (\eg backbones) is inverted by GRL,
guiding generator to deliver cross-domain invariant features so as to confuse discriminator and maximize the loss.
The overall objective of the baseline model can be formulated as
\begin{equation}
\mathcal{L} = \mathcal{L}_{det} + \lambda_1 \mathcal{L}_{da},
\end{equation}
where $\lambda_1$ is the trade-off parameter.

Following~\cite{htcn,umt}, we further interpolate the input to encourage the pixel-level consistency.
Specifically, we augment the source domain, by mixing original source images with the target-like source images generated using CycleGAN~\cite{cyclegan}. In summary, we build a very competitive baseline model with feature-level and pixel-level consistency.

\subsection{Task-specific Inconsistency Alignment}\label{alternative}

Conventional object detectors yield a \textit{single} ROI feature after FCs for both tasks of classification and localization, making it difficult to apply previous feature alignment in this \textit{coupled} space.
An intuitive way to perform task-specific alignment is to simply duplicate FCs and then align their outputs to each predictor in a DANN~\cite{dann} manner.
However, as discussed in~\cref{ablation}, such an alternative poorly decouples the task spaces and leads to insufficient alignments. More importantly, it still suffers from the lack of task-specific treatment, especially for localization task.

Following~\cite{mcd,symv1,symv2}, we propose the 
Task-specific Inconsistency Alignment to directly shrink the task-specific divergence between source and target domains. This module can be applied to both the classification and localization heads independently, as illustrated in the \textit{blue} and \textit{green} regions.
Rather than externally attaching additional discriminators, 
we use a set of auxiliary predictors to estimate the inconsistency of each domain.
By aligning them, our method can not only yield an easier approximation to domain distance, but also come up with a more natural and direct solution to perform alignment in each task space independently for detectors with multiple prediction heads.

{\bf Auxiliary predictors.} The core of our idea is employing multiple auxiliary predictors to construct an alignment mechanism between domains. Therefore, apart from the primitive classifier $C^p$ and localizer $L^p$,
two additional sets of auxiliary classifiers $C^a$ and localizers $L^a$ composed of 
$N$ classifiers $C^a_i (1 \leq i \leq N)$ and $M$ localizers $L^a_j (1 \leq j \leq M)$ respectively, are constructed on top of FCs.
To ensure a high prediction accuracy, they are all trained with labeled source data as in the primary predictors by the objective:
\begin{equation}\footnotesize
\mathcal{L}_{roi}=\frac{1}{n_s}\sum_{i=1}^{n_s}\bigg(
\sum_{j=1}^{N}\mathcal{L}^{cls}\Big(C_{j}^a\big(\hat{r}_i\big),y_i^s\Big)+\sum_{j=1}^{M}\mathcal{L}^{loc}\Big(L_{j}^{a}\big(\hat{r}_i\big),b_i^s\Big)
\bigg),
\end{equation}
where $\hat{r}_i$ represents the higher-level feature of ROI patches $r_i$ processed by FCs, 
$y_i$ and $b_i$ indicate the corresponding category label and bounding box, respectively.
For $\mathcal{L}^{cls}$ and $\mathcal{L}^{loc}$, the traditional cross-entropy and smooth-L1 losses are used.
Notably, the gradients of these auxiliary predictors are \textit{detached} when back-propagating to avoid affecting the training of primitive predictors.
In addition, to use these auxiliary predictors to perform inconsistency alignment between source and target domains, some GRLs are inserted between FCs and them to adversarially train the proposed 
task-specific inconsistency-aware losses.

\vspace{-3mm}
\subsubsection{Inconsistency Alignment Mechanism}

Previous DANN-based methods~\cite{dann} rely merely on attached binary discriminators 
to optimize task-agnostic losses. In contrast, our method optimizes fine-grained 
category- and boundary-wise multi-class losses~\cite{multiclass,bridging} for inconsistency alignment between domains,
by means of discriminators composed of various auxiliary predictors.
By nature, our objective, the \textit{alignment} to the inconsistency of auxiliary predictors' behavior (\eg decision boundaries of classifiers), essentially characterizes a more precise estimation to the \textit{margins} across domains~\cite{symv2}.
To better perceive this disagreement and perform alignment, 
we construct an integral and adversarial, single-stage training mechanism with GRL,
to cope with detectors that are too sophisticated to 
perform multi-stage iterative optimization like~\cite{mcd}.

Specifically, we initially detect the behavioral \textbf{inconsistency} of auxiliary predictors trained on the source domain
over the \textit{target} domain,
and maximize the proposed \textbf{task}-specific \textbf{i}nconsistency-\textbf{a}ware loss $\mathcal{L}_{ia}^{task}$. With GRL, the gradients back-propagated to generator (\ie FCs)
are reversed hence the loss is actually minimized for generator.
In this adversarial training, the framework reaches a dynamic \textit{equilibrium} in which the predictors are diversified to better
discriminate the discrepancy between domains, yet the generator yields sufficiently transferable features to discourage the judgments of these predictors.
In addition, the behavioral \textbf{consistency} over the \textit{source} domain of auxiliary predictors is also leveraged in a similar way.
We \textbf{maximize} the consistency-aware loss (the negative of $\mathcal{L}_{ia}^{task}$),
so as to simultaneously diversifying the source domain distribution and strengthening the predictors' capabilities.
The entire domain adaptation objective can be described as follows:
\begin{equation}\footnotesize
\begin{split}
    \mathcal{L}^{task}_{da}=&-\frac{1}{n_t}\sum_{i=1}^{n_t}\mathcal{L}_{ia}^{task}\Big(P_{1}^{a}\big(\hat{r_i}\big), P_{2}^{a}\big(\hat{r_i}\big), ...,P_{N}^{a}\big(\hat{r_i}\big)\Big)\\
    &-\frac{1}{n_s}\sum_{i=1}^{n_s}(-\mathcal{L}_{ia}^{task})\Big(P_{1}^{a}\big(\hat{r_i}\big), P_{2}^{a}\big(\hat{r_i}\big), ...,P_{N}^{a}\big(\hat{r_i}\big)\Big).
\end{split}
\end{equation}
where $task \in \{cls, loc\}$, and $P \in \{C, L\}$, the specific inconsistency measure will be explained in next subsections.

\begin{figure}
\centering
\includegraphics[width=\linewidth]{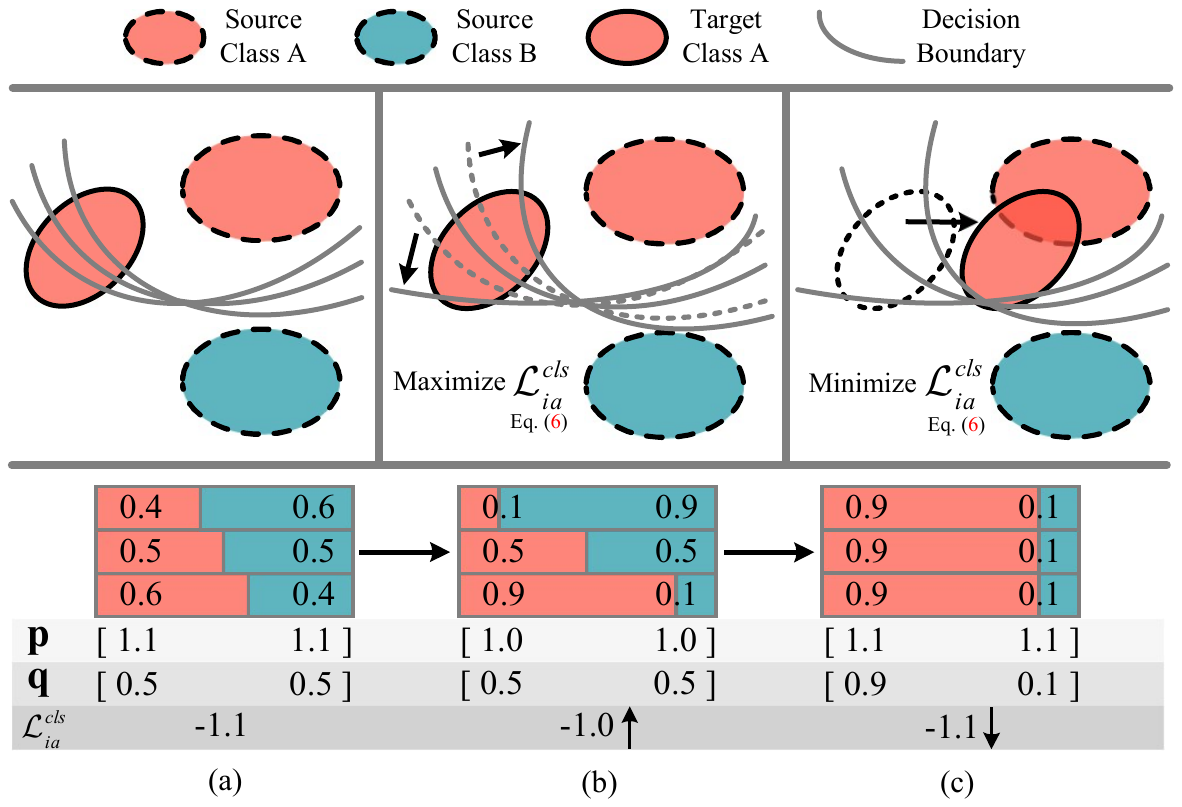}
\caption{Illustration of the effect when maximin the $\mathcal{L}_{ia}^{cls}$ over target domain in a toy example with two classes and three auxiliary classifiers. 
Best viewed in color.
(a) Initially, the behavior of classifiers is basically consistent with similar decision boundaries;
after executing a maximin optimization, we find that:
(b) the decision boundaries of the classifiers are mutually exclusive, making the probability distribution on each category sharper and the entropy lower,
hence maximizing the loss;
(c) the disparity between generated features on class A is shrunk, flattening the probability distribution and increasing the entropy,
hence minimizing the loss.}
\label{fig3}
\end{figure}

\subsubsection{Classification-specific Loss} \label{cls}

The first question is how to \textbf{capture} this behavioral disagreement among decision boundaries of auxiliary classifiers.
Different distances including L1~\cite{mcd}, Kullback-Leibler (KL)~\cite{symv2}, and Sliced Wasserstein Discrepancy (SWD)~\cite{swd} have been utilized to measure the discrepancy 
between outputs of a \textit{pair} of classifiers, but they are hard to generalize to handle multi-classifier situations.
For the score distribution constituted by auxiliary classifications on each category,
a simple assessment of its \textit{sharpness} or \textit{flatness} is expected.
Considering the stability of the optimization and also inspired by ~\cite{bnm,dirt}, 
we bound it with Shannon Entropy (SE).
Concretely, for a probability matrix $\textbf{M}\in \textbf{R}^{N\times C}$ of auxiliary predictions,
each of its column vectors $\textbf{m}_i\in \textbf{R}^{N} (1 \leq i \leq C)$ represents the predicted probabilities of all classifiers on a particular class $i$.
We can calculate an entropy vector $\textbf{p}\in \textbf{R}^{C}$ ---
each of whose elements is the entropy calculated from the corresponding softmaxed $\textbf{m}_i$
--- to describe the category-wise variations among various decision boundaries of multiple auxiliary classifiers.
Formally, the SE-driven classification-specific inconsistency-aware loss $\mathcal{L}_{ia}^{cls}$ is defined as follows:
\begin{equation}\small
\mathcal{L}_{ia}^{cls} = -\textbf{p}\cdot \textbf{q}=-\sum_{i=1}^C{\big(\sum_{j=1}^N{-\hat{m}_{ij}\log{\hat{m}_{ij}}}\big)\cdot \big(\frac{1}{N}\sum_{j=1}^N{m_{ij}}\big)},
\label{ia_cls}
\end{equation}
where $\hat{\textbf{m}}_i=\mathrm{softmax}(\textbf{m}_i)$, and $\textbf{q}$ indicates the average probability vector.
It is notable that the inner product operation between entropy vector and average probability vector is crucial,
as weighting the entropy by the confidence of distinct classes keeps the attention on the proper category.

Since the optimization of the inconsistency over target domain is our main objective, 
we take this process as an example, as depicted in~\cref{fig3}.
After solving a maximin game on $\mathcal{L}_{ia}^{cls}$, the behavior of auxiliary classifiers changes first,
and driving the probability distribution over each category to flow in a sharper and more deterministic direction.
In this case, the decision boundaries of classifiers are diversified, as shown in~\cref{fig3} (b).
Meanwhile, 
the generated target domain features are shifted towards the source domain features,
flattening the probability distribution.
In this context, features are aligned by category in the classification space,
so that greater transferability and discriminability are achieved at the same time, as illustrated in~\cref{fig3} (c).

\vspace{-3mm}
\subsubsection{Localization-specific Loss} \label{loc}

The second question lies in, how to catch the behavioral disagreement across various localizers in the \textbf{regression} space.
Unlike classification, the regression space usually exhibits a continuity and \textit{sparsity}, 
and the predicted locations are normally heterogeneously clustered in certain regions,
rendering it challenging to properly assess the dispersion of the predictions.
Some domain adaptation methods~\cite{regress,regress1} dealing with keypoint detection consider that,
shrinking the regression space by transformations contributes to alleviating the negative impact of 
sparsity on the adversarial learning of localizers.
Besides, the recent proposed \cite{gfl,gflv2}, in which the \textit{ambiguity} of multiple localizers' predictions 
on object boundaries are exploited for detecting anomalous bounding boxes, regard
top-k values along with their mean value as a robust representation to the ambiguity.

Practically, in this work, we recommend to choose the most straightforward statistic, 
the standard deviation (SD) to measure the behavioral inconsistency among auxiliary localization results.
This choice is attributed to two reasons.
First, two-stage detectors since R-CNN~\cite{rcnn} have already well constrained the regression space by \textit{linear transformations}.
Second, the L2-norm within SD is more sensitive to outliers,
which are crucial for representing the behavioral inconsistency of localizers.
The SD-driven localization-specific inconsistency-aware loss $\mathcal{L}_{ia}^{loc}$ can be formulated as
\begin{equation}\footnotesize
\mathcal{L}^{loc}_{ia} = \frac{1}{4\cdot \sqrt{M}} \sum_{i=1}^4{\left \| m_i - \frac{1}{M}\sum_{j=1}^M{m_{ij}}\right \|_2}
\label{ia_loc}
\end{equation}
where $\textbf{m}_i\in \textbf{R}^{M}$ denotes the $i$-th column vector of the prediction matrix $\textbf{M}\in \textbf{R}^{M\times 4}$ constructed by M auxiliary localizers, 
$\left \| \cdot  \right \|_2$ indicates the L2-norm. 

\begin{table*}[t]
\small
\centering
\setlength{\tabcolsep}{0.7mm}
\begin{tabular}{c|ccccccccccccccccccccc}
\toprule
Method      & aero & bcycle & bird & boat & bottle & bus  & car  & cat  & chair & cow  & table & dog & hrs  & bike & prsn & plnt & sheep & sofa & train & tv   & mAP  \\
\midrule\midrule
DAF~\cite{daf}          & 38.0 & 47.5   & 27.7 & 24.8 & 41.3 & 41.2 & 38.2 & 11.4 & 36.8  & 39.7 & 12.7  & 12.7& 31.9 & 47.8 & 55.6 & 46.3 & 12.1 & 25.6 & 51.1 & 45.5 & 34.7 \\
SWDA~\cite{swda}        & 26.2 & 48.5   & 32.6 & 33.7 & 38.5 & 54.3 & 37.1 & 18.6 & 34.8  & 58.3 & 12.5  & 12.5& 33.8 & 65.5 & 54.5 & 52.0 & 9.3  & 24.9 & 54.1 & 49.1 & 38.1 \\
SCL~\cite{scl}          & \textbf{44.7} & 50.0   & 33.6 & 27.4 & 42.2 & 55.6 & 38.3 & 19.2 & 37.9  & 69.0 & 30.1  & 26.3& 34.4 & 67.3 & 61.0 & 47.9 & 21.4 & 26.3 & 50.1 & 47.3 & 41.5 \\
HTCN~\cite{htcn}        & 33.6 & 58.9   & 34.0 & 23.4 & \textbf{45.6} & 57.0 & 39.8 & 12.0 & 39.7  & 51.3 & 20.1  & 20.1& 39.1 & 72.8 & 61.3 & 43.1 & 19.3 & \textbf{30.1} & 50.2 & \textbf{51.8} & 40.3 \\
SAP~\cite{sap} 			& 27.4 & \textbf{70.8}   & 32.0 & 27.9 & 42.4 & 63.5 & 47.5 & 14.3 & \textbf{48.2}  & 46.1 & \textbf{31.8}  & 17.9& 43.8 & 68.0 & \textbf{68.1} & 49.0 & 18.7 & 20.4 & 55.8 & 51.3 & 42.2 \\
UMT~\cite{umt}          & 39.6 & 59.1   & 32.4 & 35.0 & 45.1 & 61.9 & 48.4 & 7.5  & 46.0  & \textbf{67.6} & 21.4  & \textbf{29.5}& \textbf{48.2} & 75.9 & 70.5 & \textbf{56.7} & 25.9 & 28.9 & 39.4 & 43.6 & 44.1 \\
DBGL~\cite{dbgl}        & 28.5 & 52.3   & 34.3 & 32.8 & 38.6 & 66.4 & 38.2 & \textbf{25.3} & 39.9  & 47.4 & 23.9  & 17.9& 38.9 & 78.3 & 61.2 & 51.7 & 26.2 & 28.9 & 56.8 & 44.5 & 41.6 \\
\midrule
Source Only 			& 35.6 & 52.5   & 24.3 & 23.0 & 20.0 & 43.9 & 32.8 & 10.7 & 30.6  & 11.7 & 13.8  & 6.0 & 36.8 & 45.9 & 48.7 & 41.9 & 16.5 & 7.3  & 22.9 & 32.0 & 27.8 \\
Baseline                & 31.9 & 56.3   & 33.4 & 26.3 & 40.2 & 53.3 & 42.7 & 17.9 & 42.3  & 59.1 & 15.5  & 23.6& 35.1 & 85.2 & 63.2 & 46.3 & 22.0 & 28.4 & 51.0 & 48.2 & 41.1 \\
TIA$_{CLS}$            & 38.3 & 51.0   & \textbf{38.3} & 33.2 & 43.0 & 65.7 & 43.8 & 22.2 & 43.3  & 57.1 & 20.9  & 23.7& 38.9 & \textbf{89.4} & 64.2 & 53.8 & \textbf{38.2} & 25.0 & 52.4 & 50.5 & 44.7 \\
TIA$_{LOC}$            & 37.5 & 55.8   & 35.3 & 32.2 & \textbf{45.6} & 63.1 & 44.1 & 15.6 & 44.4  & 62.1 & 15.1  & 26.3& 38.5 & 74.3 & 65.3 & 46.9 & 30.7 & 27.2 & 55.5 & 48.9 & 43.2 \\
TIA                    & 42.2 & 66.0   & 36.9 & \textbf{37.3} & 43.7 & \textbf{71.8} & \textbf{49.7} & 18.2 & 44.9  & 58.9 & 18.2  & 29.1& 40.7 & 87.8 & 67.4 & 49.7 & 27.4 & 27.8 & \textbf{57.1} & 50.6 & \textbf{46.3} \\
\bottomrule
\end{tabular}
\caption{Experimental results (\%) of \textit{Real-to-Artistic} scenario, PASCAL VOC $\rightarrow$ Clipart.}
\label{clipart}
\end{table*}

\vspace{-3mm}
\subsubsection{Overall Objective}

Combined with the baseline model, the final objective of the proposed framework becomes
\begin{equation}
\mathcal{L} = \mathcal{L}_{det} + \lambda_1 \mathcal{L}_{da} + \lambda_2 \mathcal{L}^{cls}_{da} + \lambda_3 \mathcal{L}^{loc}_{da},
\end{equation}
where $\lambda_1$, $\lambda_2$ and $\lambda_3$ are trade-off parameters for balancing various loss components.

\subsection{Theoretical Insights}

Tracing the roots, extensive unsupervised domain adaptation methods are motivated by
the theoretical analysis in ~\cite{theory}, which states the following:

\begin{theorem}
Let $\mathcal{H}$ be the hypothesis space and let $\langle \mathcal{D}_s, f_s\rangle$ and $\langle \mathcal{D}_t, f_t\rangle$ be the two domains
consisting of a pair of distribution $\mathcal{D}$ and labeling function $f$. Hence for any $h \in \mathcal{H}$:
\begin{equation}\small
\varepsilon_{t}(h, f_t)\leq \varepsilon_{s}(h, f_s)+\frac{1}{2} d_{\mathcal{H}\Delta\mathcal{H}} (\mathcal{D}_s, \mathcal{D}_t)+\lambda ^{\ast },
\label{ben}
\end{equation}
where $\epsilon_s$ (resp. $\epsilon_t$) denotes the disagreement (\ie error) between the labeling function $f_s$ (resp. $f_t$)
and hypothesis $h$ over the source (resp. target) domain,
$d_{\mathcal{H}\Delta\mathcal{H}}$ denotes the $\mathcal{H}\Delta\mathcal{H}$ divergence between domains,
$\lambda ^{\ast }$ indicates the error of an ideal hypothesis $h^\ast$.
\end{theorem}\label{theo}

Most of existing cross-domain detectors continue the practice in DANN~\cite{dann}
and are dedicated to approximating the optimal $\mathcal{H}$-divergence (including $\mathcal{H}\Delta\mathcal{H}$-divergence) by minimizing the \textit{Jensen-Shannon} divergence~\cite{beyond}.
Then, for the two labeling functions (a classifier $f^c$ and a localizer $f^l$) possessed by all detectors, we have
\begin{equation}\small
\begin{split}
\varepsilon_{t}(h, f_t^c)\leq \varepsilon_{s}(h, f_s^c)+\frac{1}{2} d_{\mathcal{H}\Delta\mathcal{H}} (\mathcal{D}_s, \mathcal{D}_t)+\lambda ^{\ast },\\
\varepsilon_{t}(h, f_t^l)\leq \varepsilon_{s}(h, f_s^l)+\frac{1}{2} d_{\mathcal{H}\Delta\mathcal{H}} (\mathcal{D}_s, \mathcal{D}_t)+\lambda ^{\ast }.
\end{split}
\end{equation}

In this context, by narrowing a single divergence,
the target errors of both two labeling functions are restricted,
which is however, hard to be done.
Since the large differences in classification and regression spaces make it difficult for a single hypothesis 
to be consistent with both functions simultaneously,
and we also empirically found that
the target domain error of the localizer is often poorly bounded.
In response to this problem, our framework actually \textbf{decouples} the optimization of the above \textbf{divergence},
and by specifying the hypothesis on each labeling function, consistently reducing the two target errors.
Specifically, we have
\begin{equation}\small
\begin{split}
\varepsilon_{t}(h_1, f_t^c)\leq \varepsilon_{s}(h_1, f_s^c)+\frac{1}{2} d_{MCSD}^{cls} (\mathcal{D}_s, \mathcal{D}_t)+\lambda ^{\ast },\\
\varepsilon_{t}(h_2, f_t^l)\leq \varepsilon_{s}(h_2, f_s^l)+\frac{1}{2} d_{MCSD}^{loc} (\mathcal{D}_s, \mathcal{D}_t)+\lambda ^{\ast },
\end{split}
\end{equation}
where $d_{MCSD}^{cls}$ (resp. $d_{MCSD}^{loc}$) indicates the classification (resp. localization)-specific Multi-Class Scoring Disagreement~\cite{symv2} divergence, 
which is narrowed when maximining our proposed $\mathcal{L}^{cls}_{da}$ (resp. $\mathcal{L}^{loc}_{da}$).

\section{Experiments}

\subsection{Experimental Setup}

Following the default settings in ~\cite{daf,swda}, in all experiments, 
the input image is first resized to have a shorter side length of 600,
and then fed into the Faster R-CNN\cite{fasterrcnn} with ROI Align~\cite{maskrcnn}.
We train the model using the SGD optimizer with an initial learning rate of 0.001 and divide by 10 every 50k iterations.
The batch size is set to 2, one for source domain and one for target domain.
For experiments on \textit{Normal-to-Foggy} and \textit{Cross-Camera}, 
the VGG16~\cite{vgg} pretrained on ImageNet~\cite{imagenet} is employed as the detection backbone, 
and 70k iterations are trained totally.
While for \textit{Real-to-Artistic}, we use the pretrained ResNet101~\cite{resnet} instead and train a total of 120k iterations.
The numbers of auxiliary classifiers (N) and localizers (M) are set to 8 and 4, 
and the trade-off parameters $\lambda_1$, $\lambda_2$, and $\lambda_3$ are given as 1.0, 1.0 and 0.01, respectively.
We report mean average precision (mAP) with a threshold of 0.5 for evaluation.

Various state-of-the-art domain adaptive detectors are introduced for comparison, including
DAF~\cite{daf}, SWDA~\cite{swda}, MAF~\cite{maf}, SCL~\cite{scl}, HTCN~\cite{htcn},
CST~\cite{cst}, SAP~\cite{sap}, RPNPA~\cite{rpnpa}, UMT~\cite{umt}, DBGL~\cite{dbgl}, 
MeGA~\cite{mega}.
For all these methods, we cite the results in their original paper.
To verify the effectiveness of our method, we report the performance of the \textit{Baseline} model and our TIA sequentially.
We also train the Faster R-CNN using only the source images, as well as only the annotated target images,
and their performance on different scenarios is uniformly referred as \textit{Source Only}, \textit{Target Only}, respectively.

\subsection{Real to Artistic}

In this scenario, we specialize in the migration from trivial real to stylized artistic domains.
Typically, to simulate this adaptation, we use both VOC2007-trainval and VOC2012-trainval in PASCAL VOC~\cite{pascal} 
to construct natural source domain, and Clipart~\cite{dt&pl} to represent artistic target domain, according to ~\cite{dt&pl,swda,htcn}.
The Clipart shares 20 categories with PASCAL VOC, totaling 1k images, 
is employed for both training (without labels) and evaluation.

\cref{clipart} shows the results of adaptation from PASCAL VOC to Clipart.
It can be observed that our approach outperforms the previous state-of-the-arts with a notable margin (+2.2\%), achieving a mAP of 46.3\%.
Notably, the increase in localization accuracy delivers a consistent improvement over all classes,
enabling the highest mean AP with limited categories reaching the highest AP.
The overall results showcase that, a finer-grained feature alignment towards high-level abstract semantic inconsistency is essential, 
especially in such completely dissimilar scene.
Also, considering the cross-domain label shifts in the class distribution and the spatial distribution of bounding boxes, our way of shrinking both the category-wise and boundary-wise discrepancies explains the superiority of TIA.

\begin{table}[t]
\small
\centering
\setlength{\tabcolsep}{0.2mm}
\begin{tabular}{c|ccccccccc}
\toprule
Method      & bus  & bcycle & car  & cycle & person & rider & train & truck & mAP   \\ 
\midrule\midrule
DAF~\cite{daf}         & 35.3 & 27.1   & 40.5 & 20.0  & 25.0 & 31.0  & 20.2  & 22.1  & 27.6  \\
SWDA~\cite{swda}       & 36.2 & 35.3   & 43.5 & 30.0  & 29.9 & 42.3  & 32.6  & 24.5  & 34.3  \\
MAF~\cite{maf}         & 39.9 & 33.9   & 43.9 & 29.2  & 28.2 & 39.5  & 33.3  & 23.8  & 34.0  \\
SCL~\cite{scl}         & 41.8 & 36.2   & 44.8 & 33.6  & 31.6 & 44.0  & 40.7  & 30.4  & 37.9  \\
HTCN~\cite{htcn}       & 47.4 & 37.1   & 47.9 & 32.3  & 33.2 & 47.5  & 40.9  & 31.6  & 39.8  \\
CST~\cite{cst}         & 45.6 & 36.8   & 50.1 & 30.1  & 32.7 & 44.4  & 25.4  & 21.7  & 35.9  \\
SAP~\cite{sap}         & 46.8 & 40.7   & \textbf{59.8} & 30.4  & 40.8 & 46.7  & 37.5  & 24.3  & 40.9  \\
RPNPA~\cite{rpnpa}     & 43.6 & 36.8   & 50.5 & 29.7  & 33.3 & 45.6  & 42.0  & 30.4  & 39.0  \\
UMT~\cite{umt}         & \textbf{56.5} & 37.3   & 48.6 & 30.4  & 33.0 & 46.7  & 46.8  & 34.1  & 41.7  \\
MeGA~\cite{mega}       & 49.2 & 39.0   & 52.4 & 34.5  & 37.7 & \textbf{49.0}  & 46.9  & 25.4  & 41.8  \\
\midrule
Source Only            & 22.3 & 26.5   & 34.3 & 15.3  & 24.1 & 33.1  & 3.0   & 4.1   & 20.3  \\
Baseline               & 33.0 & \textbf{45.7}   & 47.9 & 33.3  & \textbf{45.5} & 36.0  & 35.0  & \textbf{37.0}  & 39.2  \\
TIA                   & 52.1 & 38.1   & 49.7 & \textbf{37.7} & 34.8 & 46.3   & \textbf{48.6}  & 31.1  & \textbf{42.3}  \\
\midrule
Target Only            & 53.1 & 36.4   & 52.8 & 36.0  & 36.2 & 46.5  & 40.2  & 34.0  & 41.9 \\
\bottomrule
\end{tabular}
\caption{Experimental results (\%) of \textit{Normal-to-Foggy} scenario, Cityscapes $\rightarrow$ Foggy Cityscapes.}
\label{city}
\end{table}

\subsection{Normal to Foggy}

The capability to accommodate various weather conditions becomes a new expectation for detectors.
In this experiment, we use Cityscapes~\cite{cityscape} and Foggy Cityscapes~\cite{foggycityscape} as the source and target domains, respectively,
to perform a transfer from regular scenes to foggy scenes.
Cityscapes comprises 3,475 images, of which 2,975 are training set and the remaining 500 are validation set.
Foggy Cityscapes is built on Cityscapes and rendered with the physical model of haze, thus both are identical in scenes and annotations.
Results are reported in the validation set of Foggy Cityscapes.

According to~\cref{city},
our proposed framework TIA obtains the highest mAP (42.3\%) over all compared methods,
and in particular, our method outperforms the \textit{Target Only} (+0.4\%) for the first time.
These results demonstrate the importance of aligning task-specific inconsistency.
Additionally, taking into account that the benchmark is close to saturation, 
the performance improvement we achieve relative to the state-of-the-art method (+0.5\%) is quite considerable.

\begin{table}[t]
\small
\centering
\setlength{\tabcolsep}{4mm}
\begin{tabular}{c|cc}
\toprule
Method      & KITTI $\rightarrow$ City & KITTI $\leftarrow$ City   \\ 
\midrule\midrule
DAF~\cite{daf}         & 38.5 & 64.1 \\
SWDA~\cite{swda}       & 37.9 & 71.0 \\
MAF~\cite{maf}         & 41.0 & 72.1 \\
SCL~\cite{scl}         & 41.9 & 72.7 \\
HTCN~\cite{htcn}       & 42.1 & 73.2 \\
CST~\cite{cst}         & 43.6 & -    \\
SAP~\cite{sap}         & 43.4 & 75.2 \\
RPNPA~\cite{rpnpa}     & -    & 75.1 \\
MeGA~\cite{mega}       & 43.0 & 75.5 \\
\midrule
Source Only            & 30.2 & 53.5 \\
Baseline               & 42.4 & 73.0 \\
TIA                   & \textbf{44.0} & \textbf{75.9} \\
\hline
\end{tabular}
\caption{Experimental results (\%) of \textit{Cross-Camera} scenario, KITTI $\leftrightarrow$ Cityscapes.}
\label{kitti}
\end{table}

\subsection{Cross Camera}
The domain gap derived from camera differences constitutes a shackle that limits applications of many deep learning algorithms. 
In this part, we adopt both KITTI~\cite{kitti} which contains 7,481 images and Cityscapes as the source and target domains
and transfer them in both adaptation directions.
In line with the protocol of ~\cite{daf}, we only evaluate detection performance on their common category, \textit{car}.

The AP on detecting cars of various adaptive detectors is reported in~\cref{kitti}.
Our method achieves the new state-of-the-art results of 44.0\% and 75.9\% in both adaptations,
also improves +1.6\% and +2.9\% respectively relative to \textit{Baseline}, manifesting once again the 
effectiveness and generalization of our approach.

\begin{figure*}
\centering
\vspace{-2mm}
\begin{minipage}[b]{.3\linewidth}
\centering
\setlength{\tabcolsep}{2mm}
\begin{tabular}{cc|c}
\toprule
Classification  & Localization    & mAP \\
\midrule
-               & -               & 41.1  \\
DANN~\cite{dann}& DANN~\cite{dann}& 42.6  \\
L1              & L1              & 43.2  \\ 
KL              & L1              & 43.7  \\ 
SWD~\cite{swd}  & SWD~\cite{swd}  & 44.4  \\ 
$\mathcal{L}_{ia}^{cls}$ (\ref{ia_cls}) & $\mathcal{L}_{ia}^{loc}$ (\ref{ia_loc}) & 46.3  \\
\hline
\end{tabular}
\captionof{table}{Ablation study on the effect on subtasks.}
\label{tab:ablation}
\end{minipage}
\hspace{3mm}
\begin{minipage}[b]{.3\linewidth}
\centering
\small
\includegraphics[width=\linewidth]{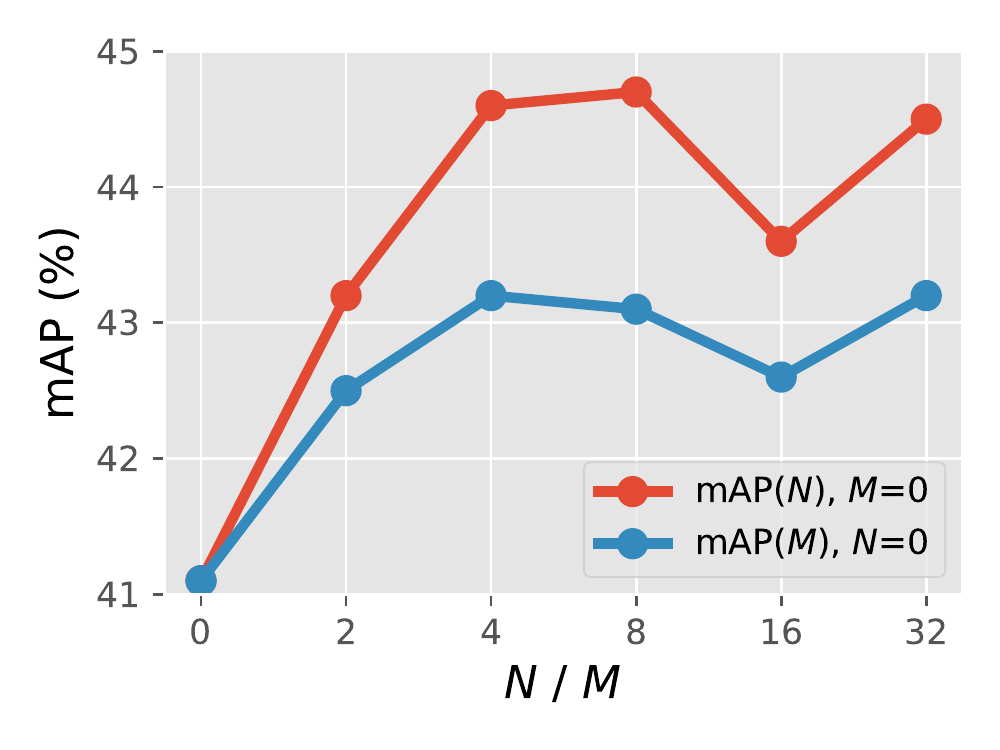}
\vspace{-7mm}
\caption{Ablation study on the
effect of number of auxiliary predictors $N$, $M$.
}
\label{fig5}
\end{minipage}
\hspace{3mm}
\begin{minipage}[b]{.33\linewidth}
\centering
\small
\includegraphics[width=\linewidth]{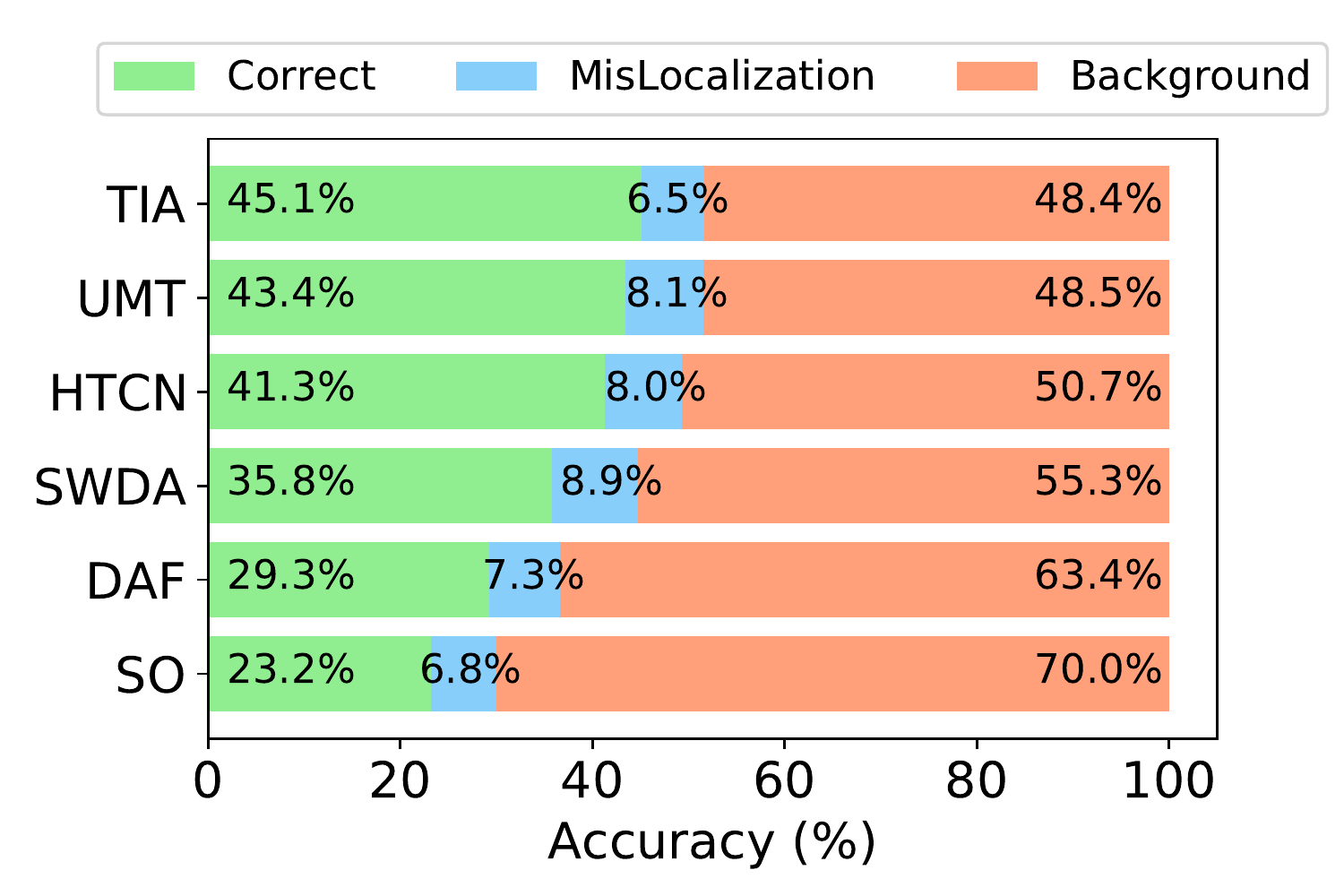}
\vspace{-7mm}
\caption{Error analysis of highest confident detections.
(SO refers to \textit{Source Only})}
\label{fig4}
\end{minipage}
\end{figure*}

\section{Analysis}

\subsection{Ablation Study} \label{ablation}

\textbf{Effect on subtasks.} 
\cref{clipart} also demonstrates the effectiveness of TIA on both subtasks of classification and localization.
As represented by \textit{TIA$_{CLS}$} and \textit{TIA$_{LOC}$}, our proposed classification- and localization-specific inconsistency alignment bring consistent improvements (+3.6\% and +2.1\%).
These results indicate that aligning the inconsistency in each task space is effective in enhancing both the category-wise and boundary-wise transferability.
Moreover, because of the orthogonality of classification and localization, and
boundary-wise alignment always leads to congruent improvement for the detector,
revealing the significance of learning a cross-domain localizer.

\textbf{Effect of inconsistency losses.}
The superiority of our proposed $\mathcal{L}_{ia}^{cls}$ (\ref{ia_cls}) and $\mathcal{L}_{ia}^{loc}$ (\ref{ia_loc})
is validated by~\cref{tab:ablation}, on the PASCAL VOC $\rightarrow$ Clipart benchmark.
For compared models from third to fifth rows, the number of auxiliary classifiers ($N$) and localizers ($M$) is fixed to 2.
In comparison to the practical L1 loss used in \textit{MCD}~\cite{mcd}, KL divergence used in~\cite{symv2},
the recently proposed \textit{SWD}~\cite{swd}, and the intuitive alternative as mentioned in~\cref{alternative},
our losses better capture the disagreement and ambiguity of the behavior in auxiliary classifiers and localizers,
thus suggesting more precise measures to cross-domain discrepancies.
In particular, the use of L1 loss on our framework (shown in the third row) can be regarded as a suitable alternative to MCD~\cite{mcd}, since experimentally, it fails to work on detection. Therefore, the improvement of our TIA \textit{\wrt} it reveals both the stability of our optimization and the superiority of the proposed losses.
Additionally, the way of simply repeating FCs and then aligning their output features to each predictor
based on \textit{DANN}~\cite{dann} (shown in the second row) brings a limited performance gain.
This is reasonable since it actually shares substantial layers before FCs, resulting in poor decoupling of the feature space and inadequate task-specific alignment. Meanwhile, it also lacks task-specific treatment and hence still suffers from the rising of localization error in the classification space of the domain discriminator.

\textbf{Effect of number of auxiliary predictors $N$, $M$.}
To reveal more clearly the impact of number of auxiliary predictors, \ie $N$ or $M$, on their accordingly tasks, we fix one to 0 and exponentially vary the other from 0 to 32, and~\cref{fig5} depicts their impact.
Obviously, $N$ contributes more to the overall performance than $M$,
suggesting a rational representation of category-wise differences is of greater significance.
In addition, in sparse regression spaces, the growth in the number of localizers is of limited benefit in capturing the inconsistency. And we speculate that this is because the localization results are typically heterogeneously clustered in certain regions.

\subsection{Error Analysis}

To demonstrate that our framework is capable of promoting 
the discriminability of features towards both classification and localization tasks,
we analyze the accuracies of \textit{Source Only}~\cite{fasterrcnn}, \textit{DAF}~\cite{daf}, \textit{SWDA}~\cite{swda}, \textit{HTCN}~\cite{htcn}, \textit{UMT}~\cite{umt} and our TIA
caused by most confident detections on the Foggy Cityscapes $\rightarrow$ Cityscapes task.
In line with ~\cite{daf}, we categorize the detections into 3 types: 
\textbf{Correct} (IoU with GT $\geq$ 0.5), \textbf{MisLocalization} (0.5 \textgreater IoU with GT $\geq$ 0.3) and \textbf{Background} (IoU with GT \textless 0.3). 
For each class, we select top-K predictions where K is the number of ground-truth bounding boxes in this class,
and report the mean percentage of each type across all categories.
As shown in~\cref{fig4}, in comparison with previous mainstream cross-domain detectors,
not only does our TIA clearly improve the number of correct detections (\textit{green} color) and reduce the number of false positives,
but more importantly, it lowers mislocalization error relative to \textit{Source Only} for the first time (\textit{blue} color).
This proves that our TIA boosts the transferability of features while also enhancing their awareness towards both tasks, especially the localization task.

\vspace{-1mm}
\section{Conclusions and Limitations}
\vspace{-1mm}

In this paper, we propose a new method TIA, by developing fine-grained feature alignment in separate task spaces in terms of inconsistency, sufficiently strengthening both the classification and localization capabilities of the detector.
Extensive experiments demonstrate the effectiveness of TIA.
Nevertheless, the issues of label shift and training stability inherent in domain adaptation still limit TIA, and research in these regards will become future work.

\vspace{-1mm}
\section*{Acknowledgments}
\vspace{-1mm}

This work is supported by National Natural Science Foundation of China (No.62076119, No.61921006), Program for Innovative Talents and Entrepreneur in Jiangsu Province, and Collaborative Innovation Center of Novel Software Technology and Industrialization.

{\small
\bibliographystyle{ieee_fullname}
\bibliography{main}
}

\clearpage
\appendix
\section*{Appendix}

\section{More Implementation Details}

\begin{table*}[t]
\centering
\small
\begin{minipage}{0.30\linewidth}
  \centering
  \begin{tabular}{|c|}
    \hline
    Discriminator $D_1$ \\
    \hline
    Conv 1 × 1 × 256, stride 1, pad 0 \\
    ReLU \\
    Conv 1 × 1 × 128, stride 1, pad 0 \\
    ReLU \\
    Conv 1 × 1 × 1, stride 1, pad 0 \\
    Sigmoid \\
    \hline
  \end{tabular}
\end{minipage}
\begin{minipage}{0.30\linewidth}
  \centering
  \begin{tabular}{|c|}
    \hline
    Discriminator $D_2$ and $D_3$ \\
    \hline
    Conv 3 × 3 × 512, stride 2, pad 1 \\
    Batch Normalization, ReLU, Dropout \\
    Conv 3 × 3 × 128, stride 2, pad 1 \\
    Batch Normalization, ReLU, Dropout \\
    Conv 3 × 3 × 128, stride 2, pad 1 \\
    Batch Normalization, ReLU, Dropout \\
    Average Pooling \\
    Fully connected 128 × 2 \\
    \hline
  \end{tabular}
\end{minipage}
\begin{minipage}{0.30\linewidth}
  \centering
  \begin{tabular}{|c|}
    \hline
    Discriminator $D_4$  \\
    \hline
    Conv 3 × 3 × 512, stride 2, pad 1 \\
    ReLU \\
    Conv 3 × 3 × 128, stride 2, pad 1 \\
    ReLU \\
    Conv 3 × 3 × 128, stride 2, pad 1 \\
    ReLU \\
    Average Pooling \\
    Fully connected 128 × 2 \\
    \hline
  \end{tabular}
\end{minipage}
\caption{Architecture of discriminators.}
\label{disc}
\end{table*}

In this section, we present more details about the \textit{Baseline} model.
As mentioned in main text (Sec. 3.1), for higher semantic consistency,
we adhere to the mainstream practice of aligning features on the source and target domains,
at both mid-to-upper layers of the backbone (\ie image-level) and ROI layer (\ie instance-level),
with the help of Gradient Reversal Layer (GRL)~\cite{dann}.
Concretely, in consistent with~\cite{scl},
for the features output from the last three blocks of VGG16~\cite{vgg}, or last three layers of ResNet101~\cite{resnet},
we feed them into separate discriminators 
($D_1$, $D_2$ and $D_3$, their concrete architecture is shown in~\cref{disc}) 
connected via a GRL to determine the domain to which the features belong. 
After that, three image-level domain adaptation losses are calculated as follows:
\begin{equation}
\begin{split}
    \mathcal{L}_{da}^{img1} &= \frac{1}{n_s \cdot H\cdot W}\sum_{i=1}^{n_s}\sum_{w=1}^{W}\sum_{h=1}^{H}{D_1(x_i)^2_{wh}} \\
    &+ \frac{1}{n_t \cdot H\cdot W}\sum_{i=1}^{n_t}\sum_{w=1}^{W}\sum_{h=1}^{H}{(1-D_1(x_i)_{wh})^2},
\end{split}
\end{equation}
\begin{equation}
\begin{split}
    \mathcal{L}_{da}^{img2} &= \frac{1}{n_s}\sum_{i=1}^{n_s}{\mathcal{L}_{ce}(D_2(x_i^{'}), d_i^s)} \\
    &+ \frac{1}{n_t}\sum_{i=1}^{n_t}{\mathcal{L}_{ce}(D_2(x_i^{'}), d_i^t)},
\end{split}
\end{equation}
\begin{equation}
\begin{split}
    \mathcal{L}_{da}^{img3} &= \frac{1}{n_s}\sum_{i=1}^{n_s}{\mathcal{L}_{fl}(D_3(x_i^{''}), d_i^s)} \\
    &+ \frac{1}{n_t}\sum_{i=1}^{n_t}{\mathcal{L}_{fl}(D_2(x_i^{''}), d_i^t)},
\end{split}
\end{equation}
where $x_i$, $x_i^{'}$ and $x_i^{'''}$ denotes the features output from the last three blocks of the backbone
for the $i$-th training image, $d_i$ indicates the corresponding domain label,
and $n_s$ and $n_t$ refer to the total number of images within a mini-batch 
in source and target domains, respectively.
Besides, ${L}_{ce}$ suggests the cross-entropy loss, 
while the ${L}_{fl}$ indicates the focal loss, with its $\gamma$ set to 5 following ~\cite{swda}.
Likewise, the alignment of high-level feature patches (ROIs) is also employed.
With the discriminator $D_4$ illustrated in~\cref{disc}, the instance-level loss is formally as
\begin{equation}
\begin{split}
    \mathcal{L}_{da}^{ins}&= \frac{1}{n_s}\sum_{i=1}^{n_s}\mathcal{L}_{ins}(D(r_i),d_i^s)\\
    &+\frac{1}{n_t}\sum_{i=1}^{n_t}\mathcal{L}_{ins}(D(r_i),d_i^t),    
\end{split}
\end{equation}
where $r_i$ denotes the $i$-th ROI and $d_i$ indicates the corresponding domain label.
As for $\mathcal{L}_{ins}$, we use cross-entropy loss for the \textit{Normal-to-Foggy} and \textit{Cross-Camera} scenarios
and focal loss for the \textit{Real-to-Artistic} scenario, with $\gamma$ being also set to 5.

In conclusion, the overall training objective of \textit{Baseline} becomes:
\begin{equation}
\mathcal{L} = \mathcal{L}_{det} +\lambda_1 (\mathcal{L}_{da}^{img1} +\mathcal{L}_{da}^{img2} +\mathcal{L}_{da}^{img3} + \mathcal{L}_{da}^{ins}),
\end{equation}
where $\lambda_1$ is set to 1.0.
Additionally, we concatenate the image-level features processed by previous three discriminators
with the high-level ROI representation after FCs, in a manner similar to ~\cite{swda,scl},
to realize greater training stability.

\section{Additional Ablation Study}

For the localization-specific inconsistency alignment module, 
the effect of \textbf{different measures of dispersion} is further investigated here.
To reveal more clearly their impact on the localization branch,
we remove the classification branch.
The results on the \textit{Real-to-Artistic} scenario are displayed in~\cref{abl_}.
It showcases that
(1) a measure that is closer to the original scale is preferred;
(2) L2-norm delivers a more appropriate and precise estimate to behavioral uncertainty among diverse localizers.

\begin{table}[t]
\small
\centering
\setlength{\tabcolsep}{4mm}
\begin{tabular}{c|c}
\toprule
Measurement      & mAP   \\ 
\midrule
Mean absolute deviation & 42.7 \\
Variance                & 41.8 \\
Standard deviation      & 43.2 \\
\bottomrule
\end{tabular}
\caption{Ablation study on different measures of dispersion.}
\label{abl_}
\end{table}

\section{Visualization}

\begin{figure*}
\centering

\begin{subfigure}{.32\linewidth}
  \centering
  \includegraphics[width=0.7\linewidth]{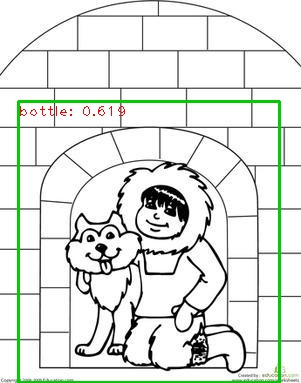}
\end{subfigure}
\begin{subfigure}{.32\linewidth}
  \centering
  \includegraphics[width=0.7\linewidth]{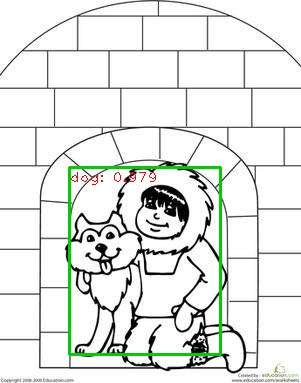}
\end{subfigure}
\begin{subfigure}{.32\linewidth}
  \centering
  \includegraphics[width=0.7\linewidth]{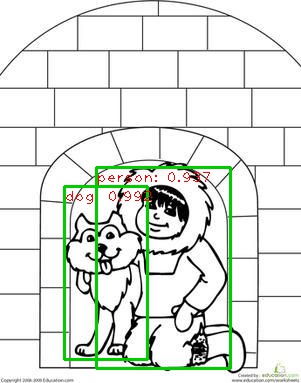}
\end{subfigure}
\\[\smallskipamount]

\begin{subfigure}{.32\linewidth}
  \centering
  \includegraphics[width=\linewidth]{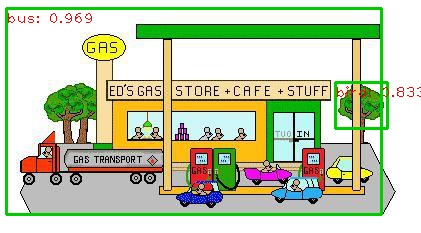}
\end{subfigure}
\begin{subfigure}{.32\linewidth}
  \centering
  \includegraphics[width=\linewidth]{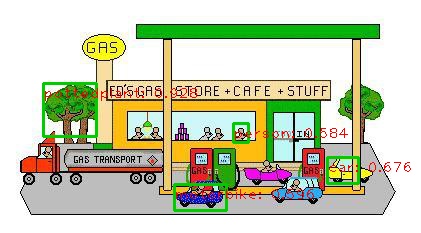}
\end{subfigure}
\begin{subfigure}{.32\linewidth}
  \centering
  \includegraphics[width=\linewidth]{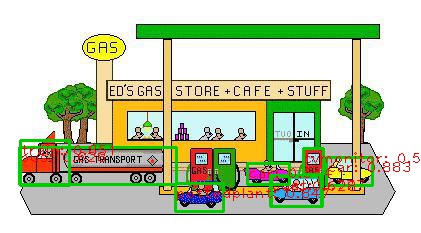}
\end{subfigure}
\\[\smallskipamount]

\begin{subfigure}{.32\linewidth}
  \centering
  \includegraphics[width=\linewidth]{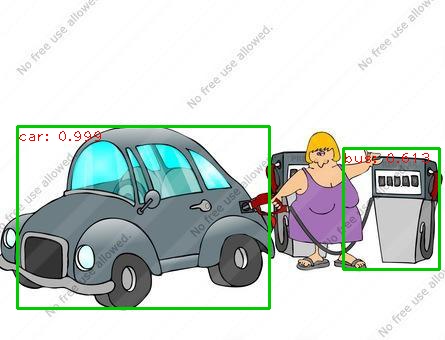}
\end{subfigure}
\begin{subfigure}{.32\linewidth}
  \centering
  \includegraphics[width=\linewidth]{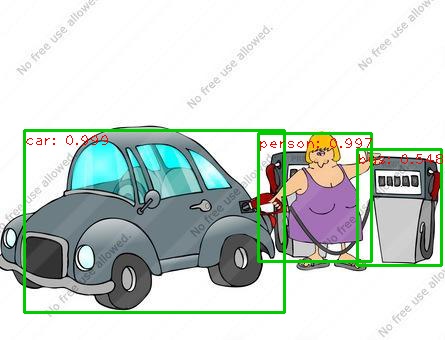}
\end{subfigure}
\begin{subfigure}{.32\linewidth}
  \centering
  \includegraphics[width=\linewidth]{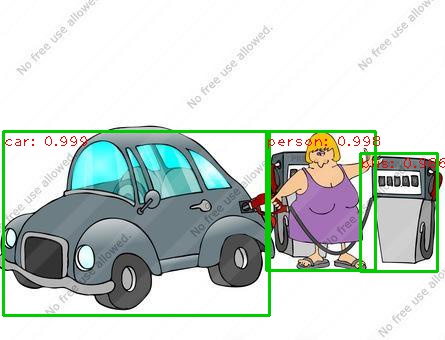}
\end{subfigure}
\\[\smallskipamount]

\begin{subfigure}{.32\linewidth}
  \centering
  \includegraphics[width=\linewidth]{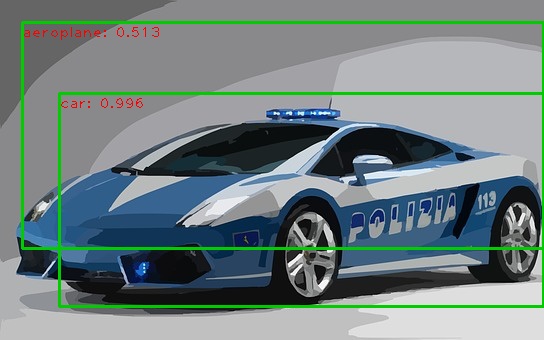}
\end{subfigure}
\begin{subfigure}{.32\linewidth}
  \centering
  \includegraphics[width=\linewidth]{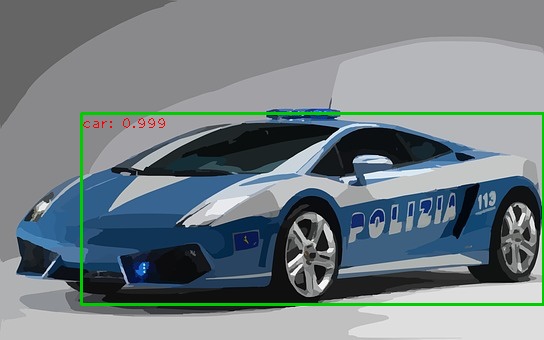}
\end{subfigure}
\begin{subfigure}{.32\linewidth}
  \centering
  \includegraphics[width=\linewidth]{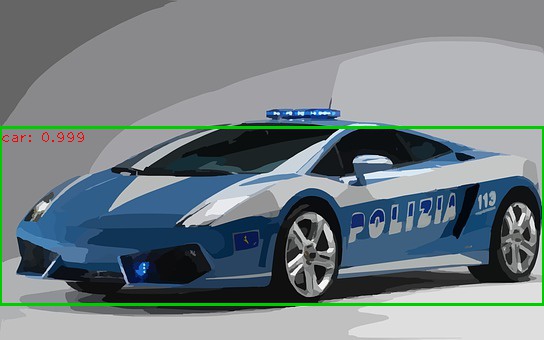}
\end{subfigure}
\\[\smallskipamount]

\begin{subfigure}{.32\linewidth}
  \centering
  \includegraphics[width=\linewidth]{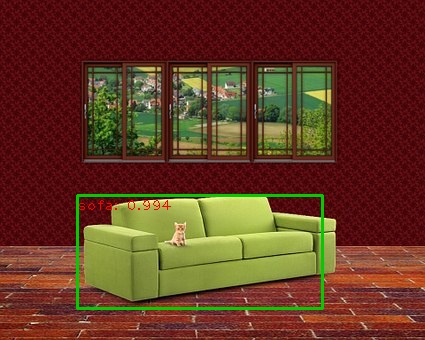}
  \caption{Source Only}
\end{subfigure}
\begin{subfigure}{.32\linewidth}
  \centering
  \includegraphics[width=\linewidth]{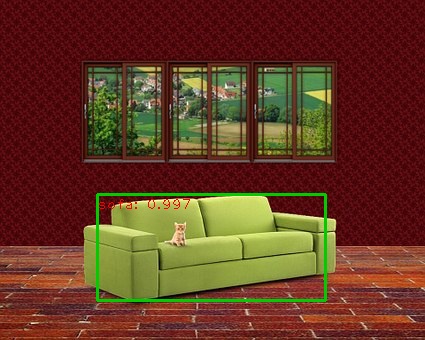}
  \caption{UMT~\cite{umt}}
\end{subfigure}
\begin{subfigure}{.32\linewidth}
  \centering
  \includegraphics[width=\linewidth]{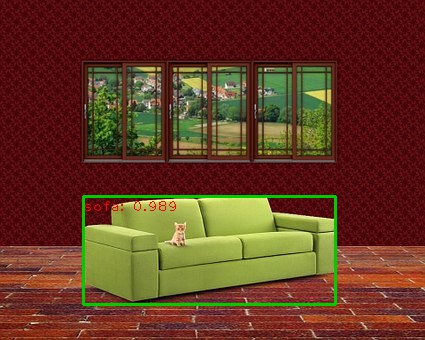}
  \caption{Our TIA}
\end{subfigure}
\\[\smallskipamount]

\caption{
Illustration of the detection results on the PASCAL VOC $\rightarrow$ Clipart benchmark.
Compared to \textit{Source Only}, \textit{UMT}'s localization performance is worse, while ours is better.
}
\label{vis1}
\end{figure*}

\begin{figure*}
\small
\centering
\begin{subfigure}{.32\linewidth}
  \centering
  \includegraphics[width=\linewidth]{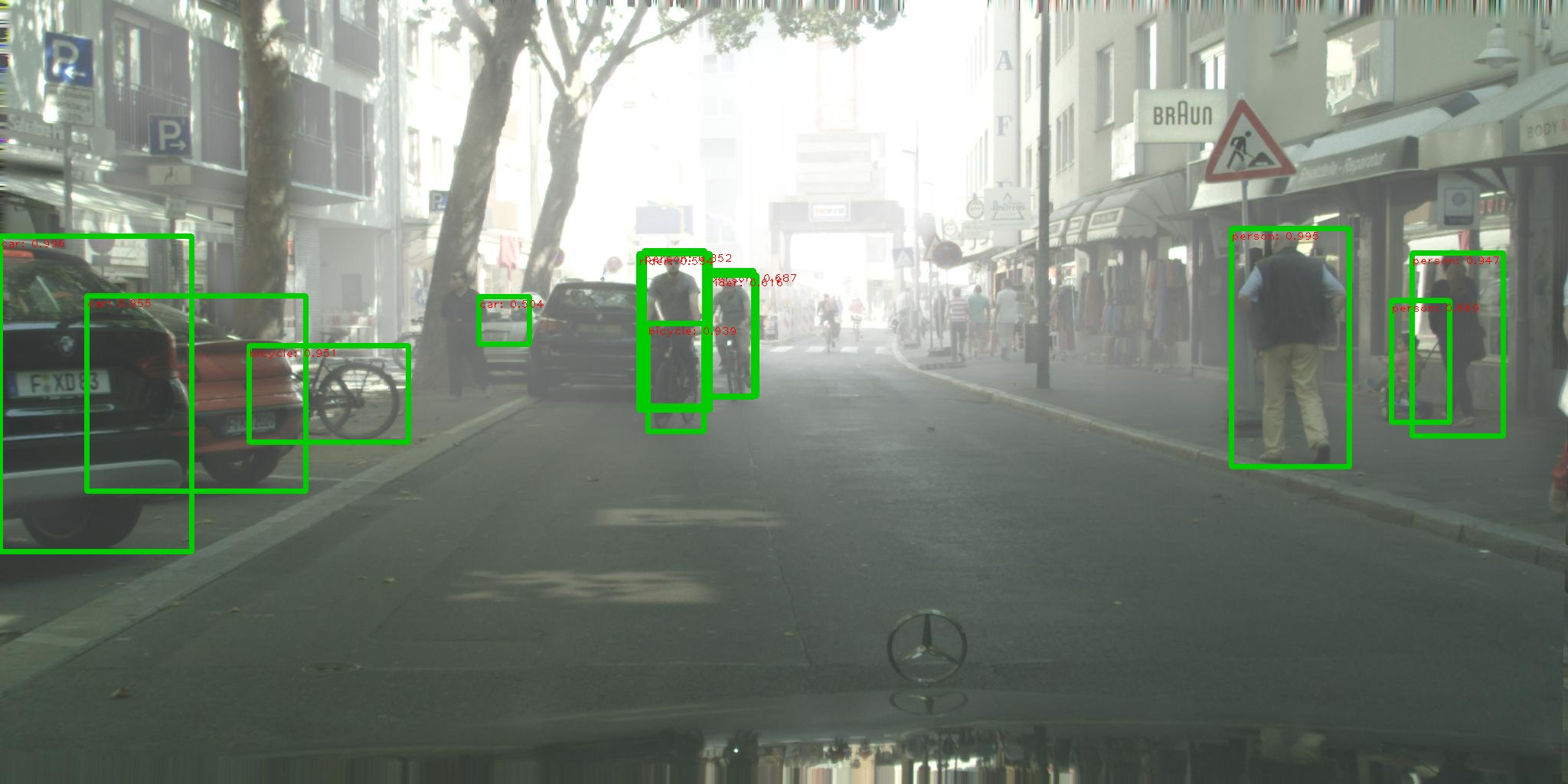}
\end{subfigure}
\begin{subfigure}{.32\linewidth}
  \centering
  \includegraphics[width=\linewidth]{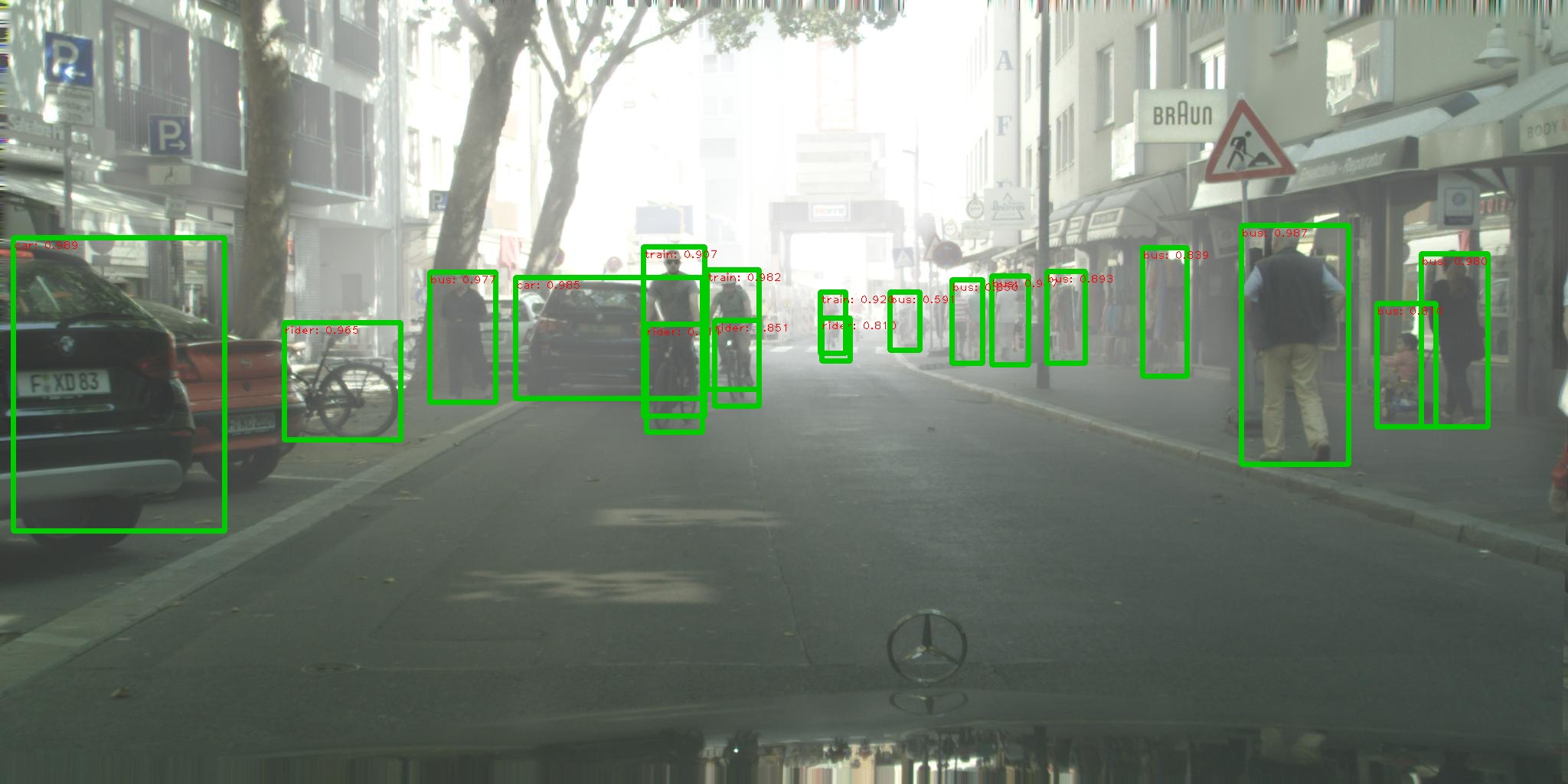}
\end{subfigure}
\begin{subfigure}{.32\linewidth}
  \centering
  \includegraphics[width=\linewidth]{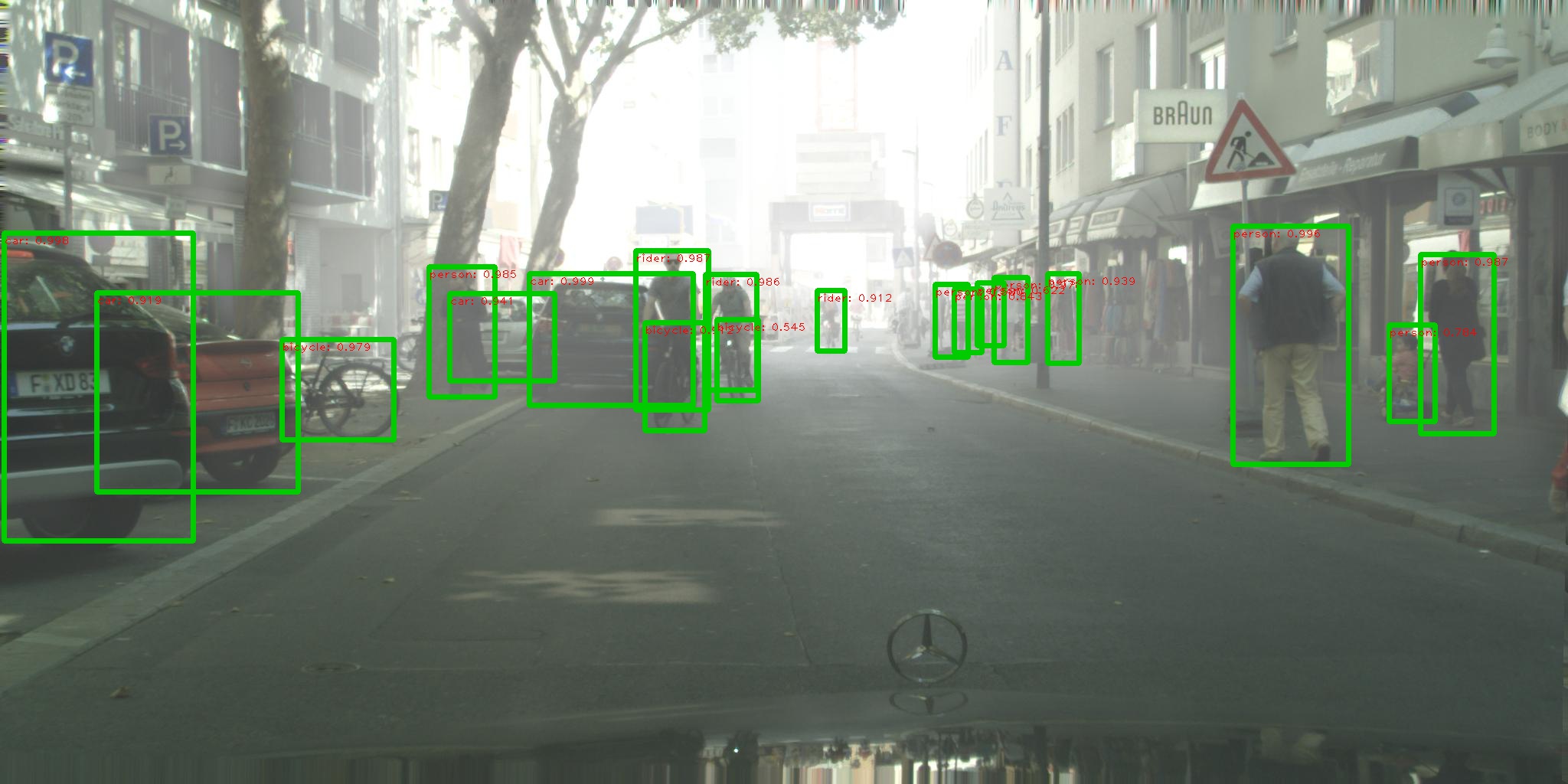}
\end{subfigure}
\\[\smallskipamount]

\begin{subfigure}{.32\linewidth}
  \centering
  \includegraphics[width=\linewidth]{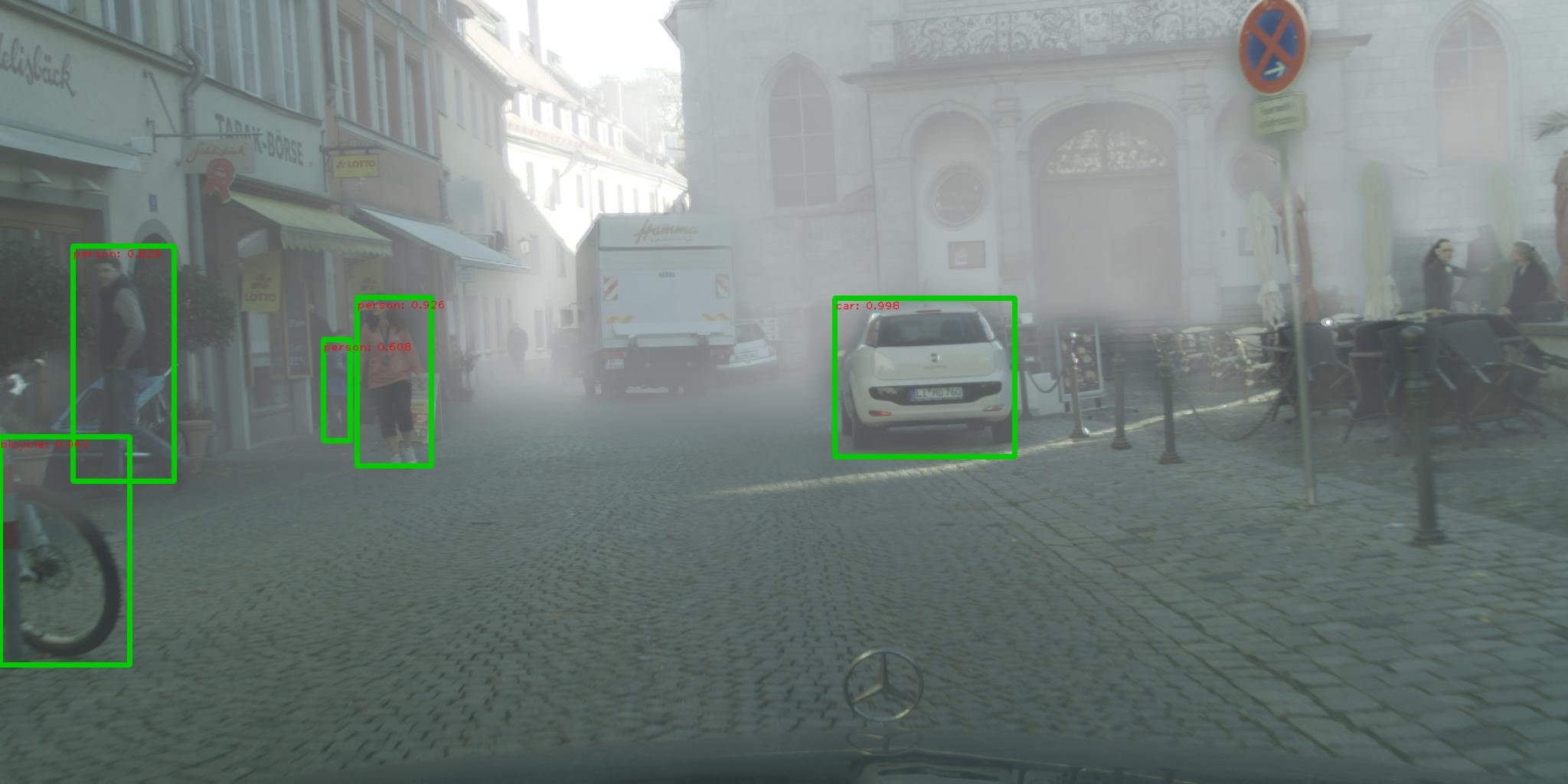}
\end{subfigure}
\begin{subfigure}{.32\linewidth}
  \centering
  \includegraphics[width=\linewidth]{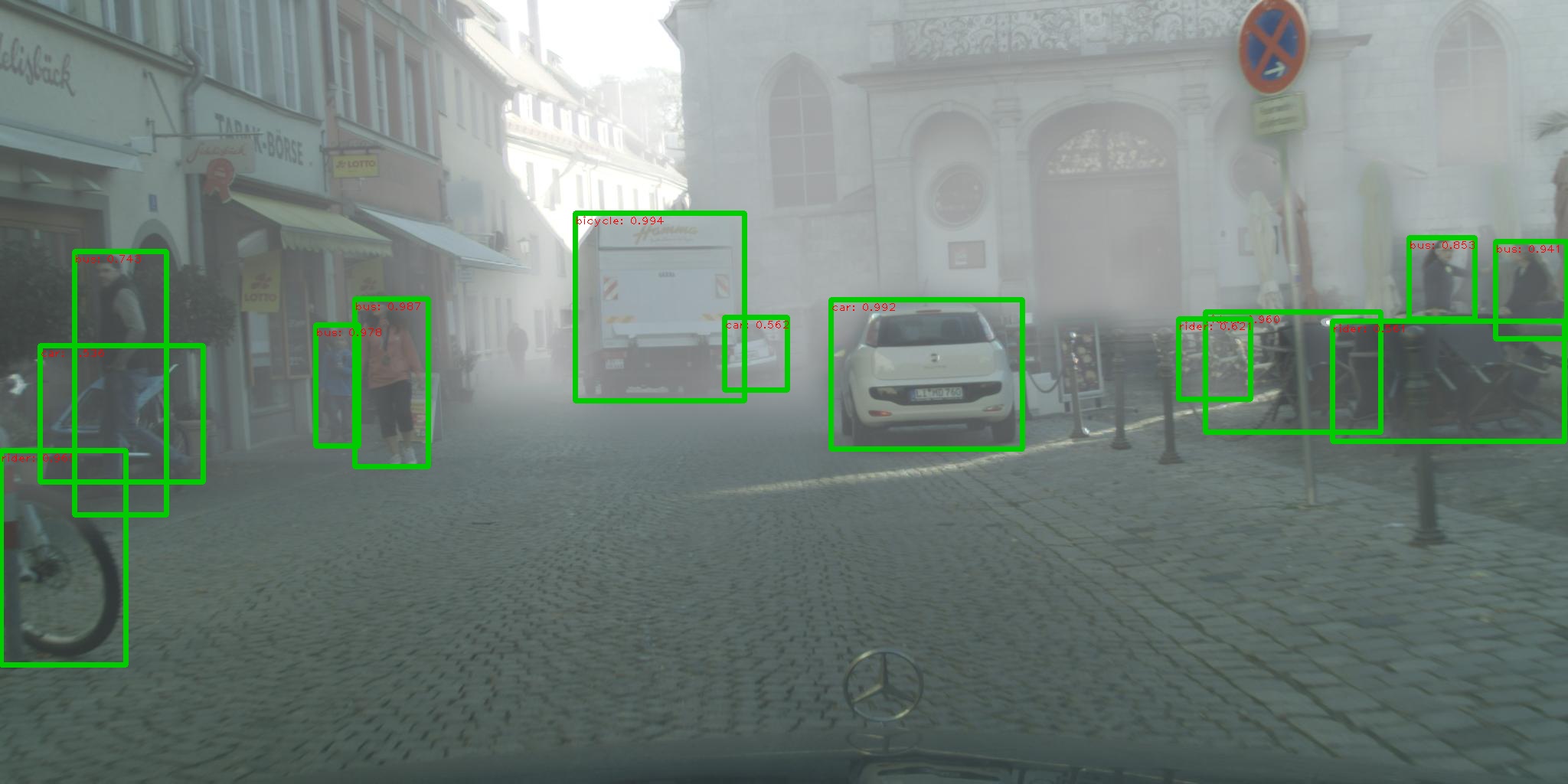}
\end{subfigure}
\begin{subfigure}{.32\linewidth}
  \centering
  \includegraphics[width=\linewidth]{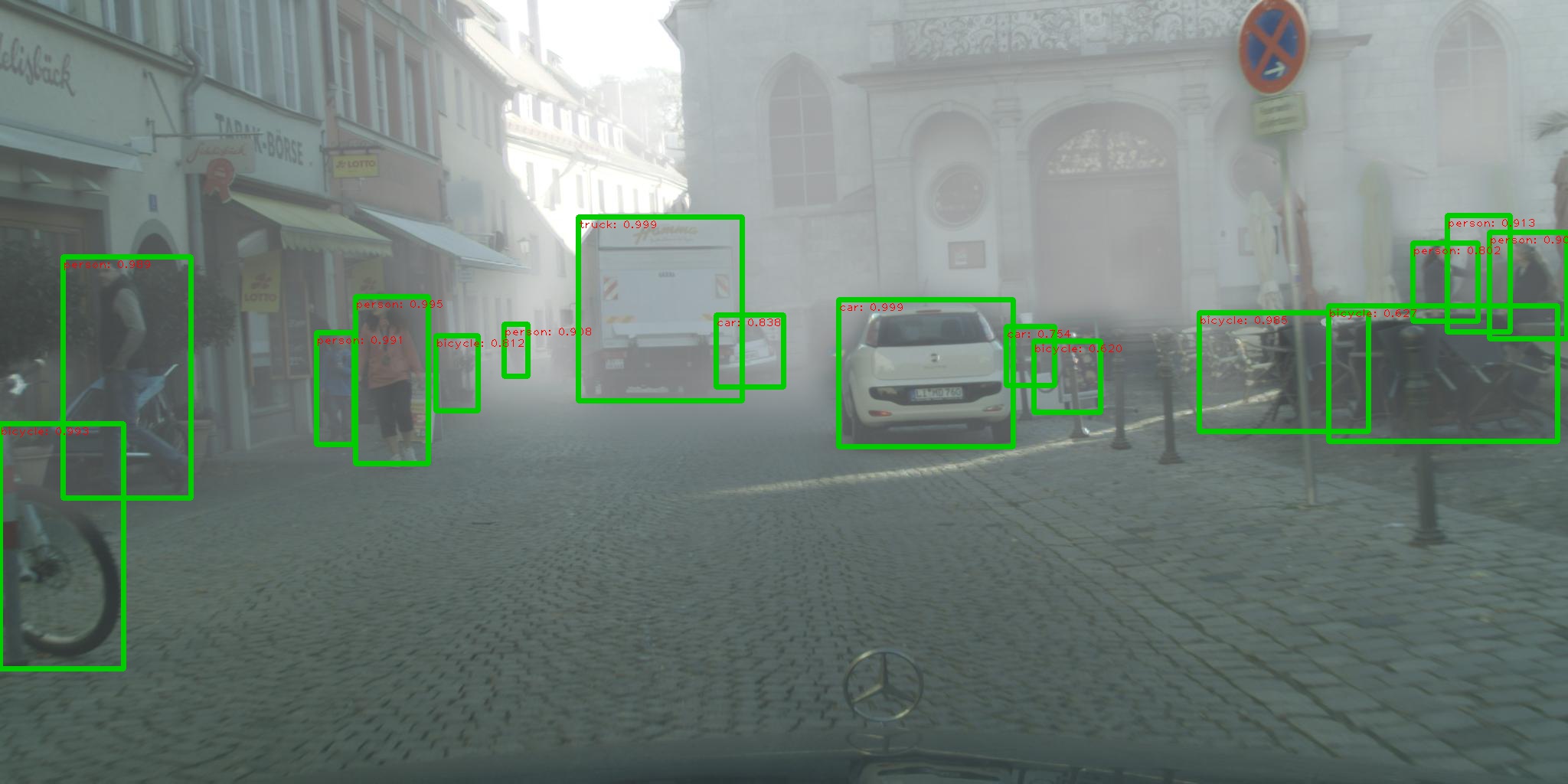}
\end{subfigure}
\\[\smallskipamount]

\begin{subfigure}{.32\linewidth}
  \centering
  \includegraphics[width=\linewidth]{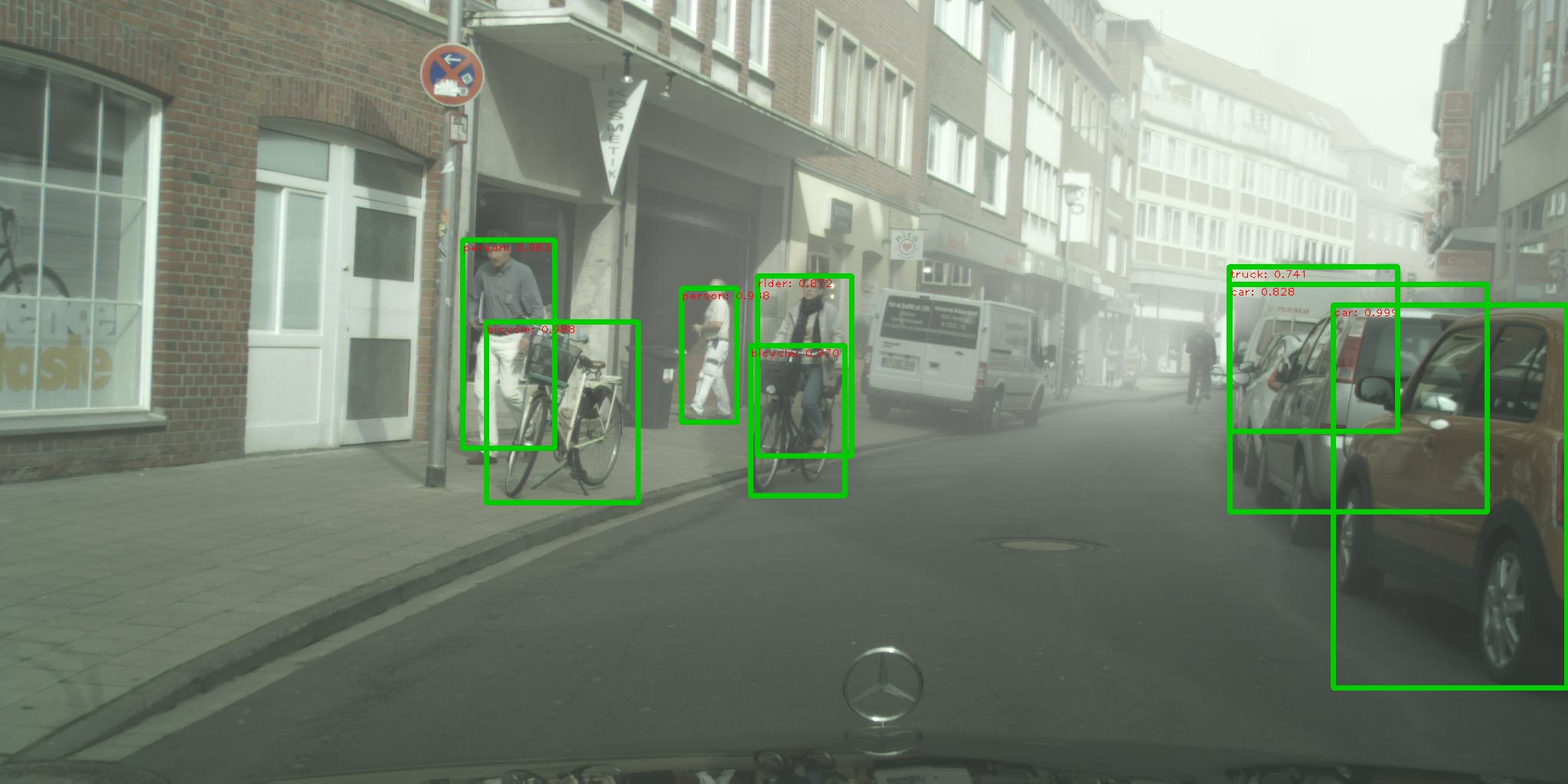}
\end{subfigure}
\begin{subfigure}{.32\linewidth}
  \centering
  \includegraphics[width=\linewidth]{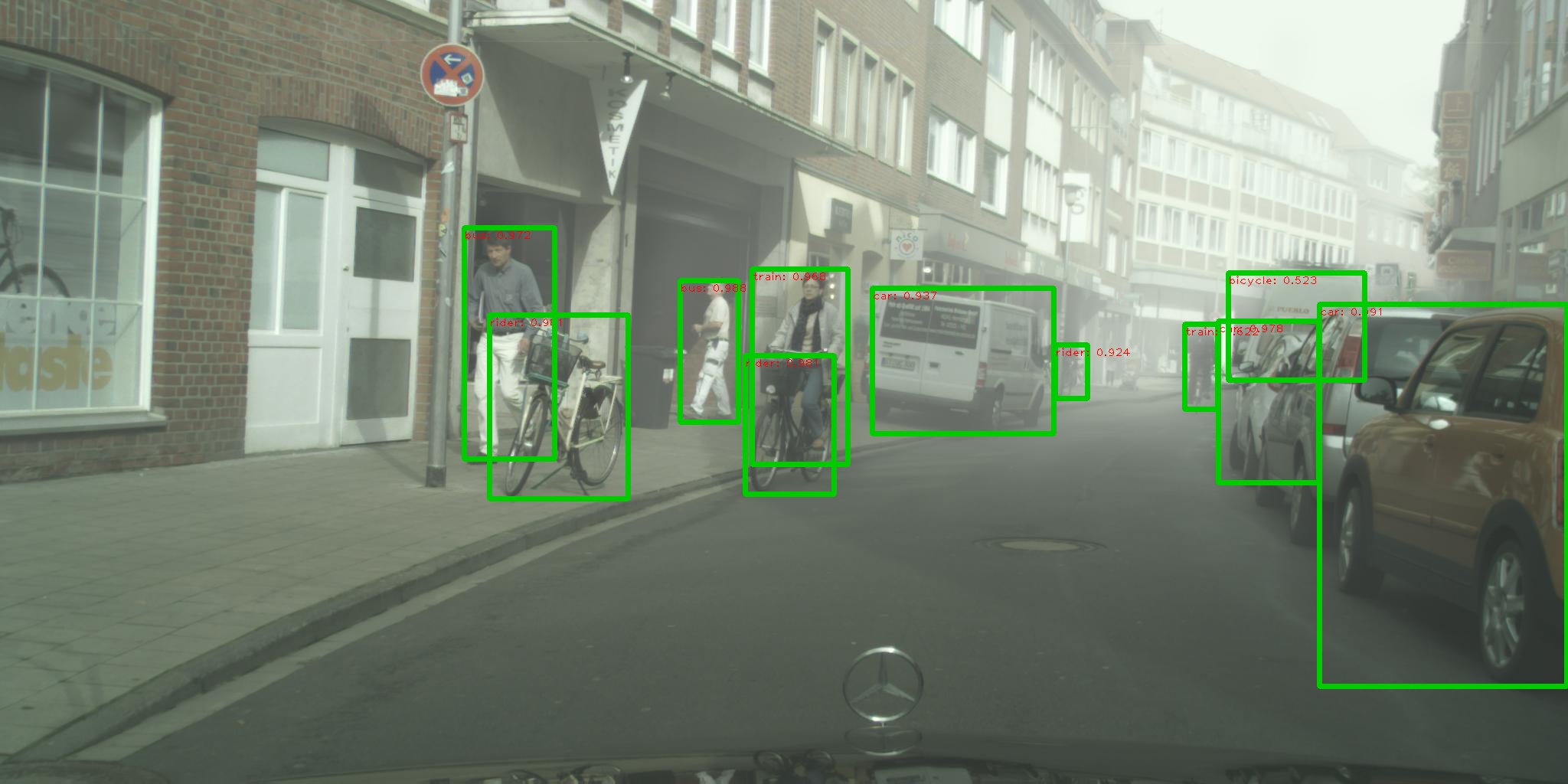}
\end{subfigure}
\begin{subfigure}{.32\linewidth}
  \centering
  \includegraphics[width=\linewidth]{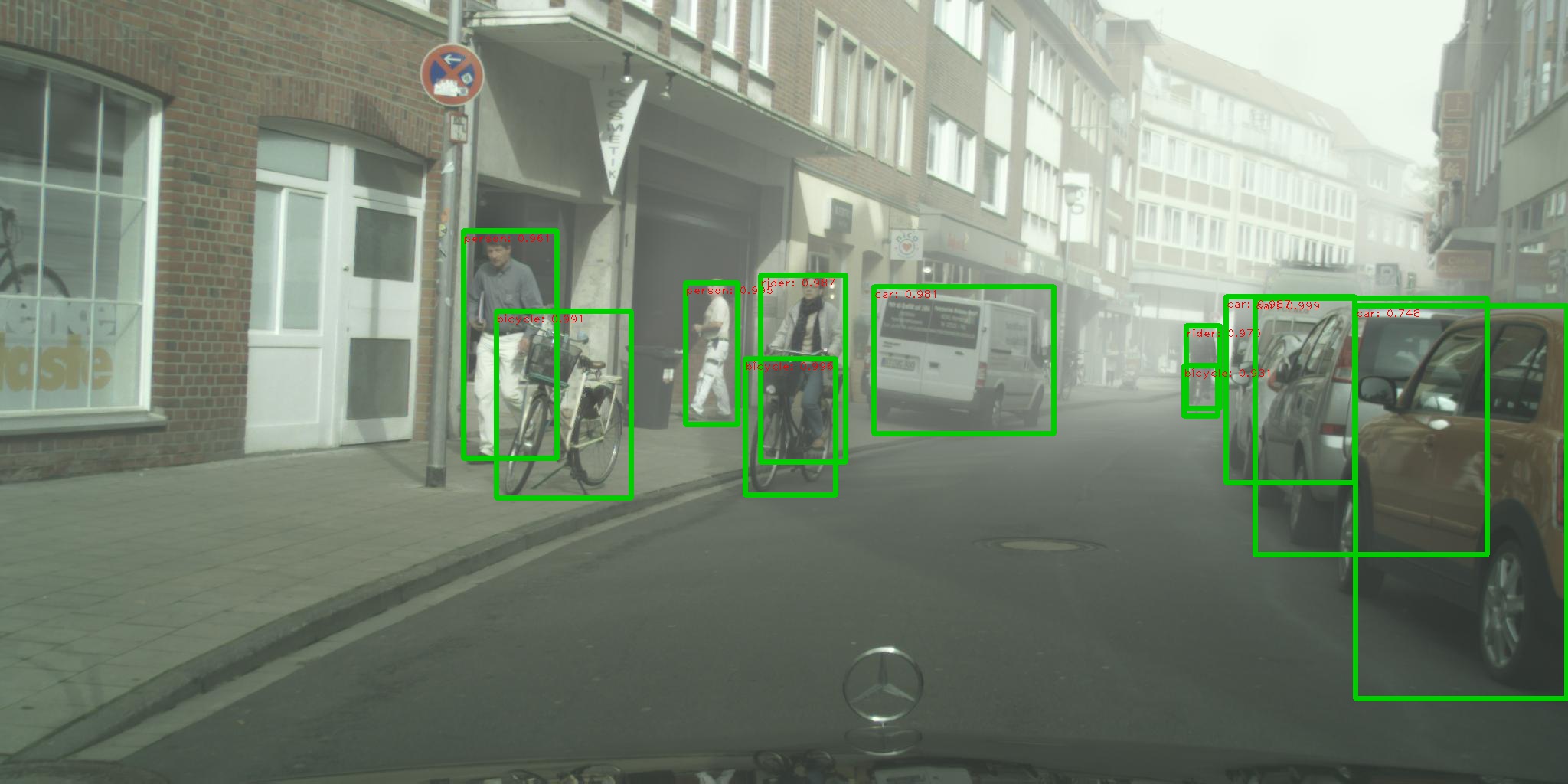}
\end{subfigure}
\\[\smallskipamount]

\begin{subfigure}{.32\linewidth}
  \centering
  \includegraphics[width=\linewidth]{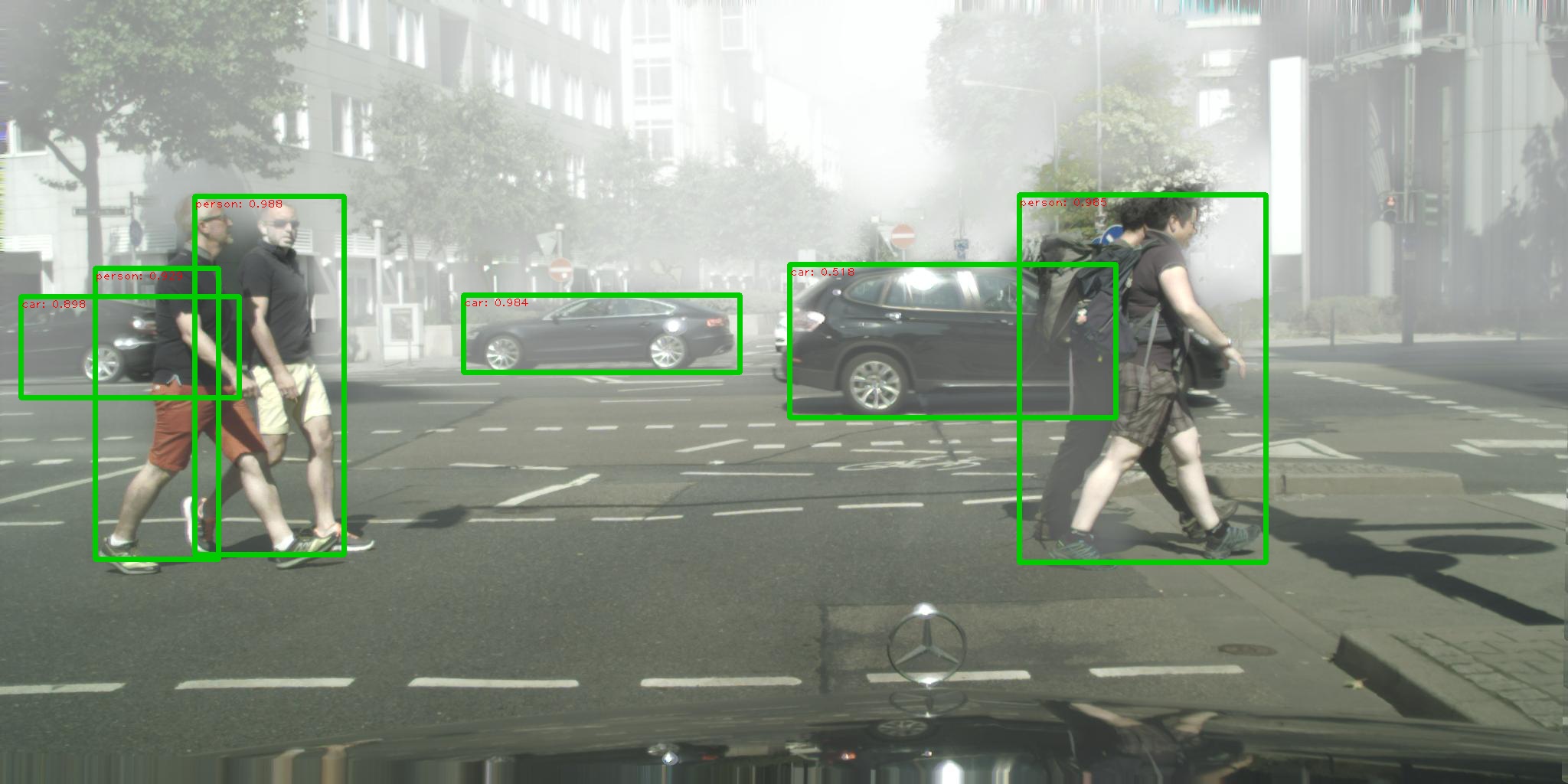}
\end{subfigure}
\begin{subfigure}{.32\linewidth}
  \centering
  \includegraphics[width=\linewidth]{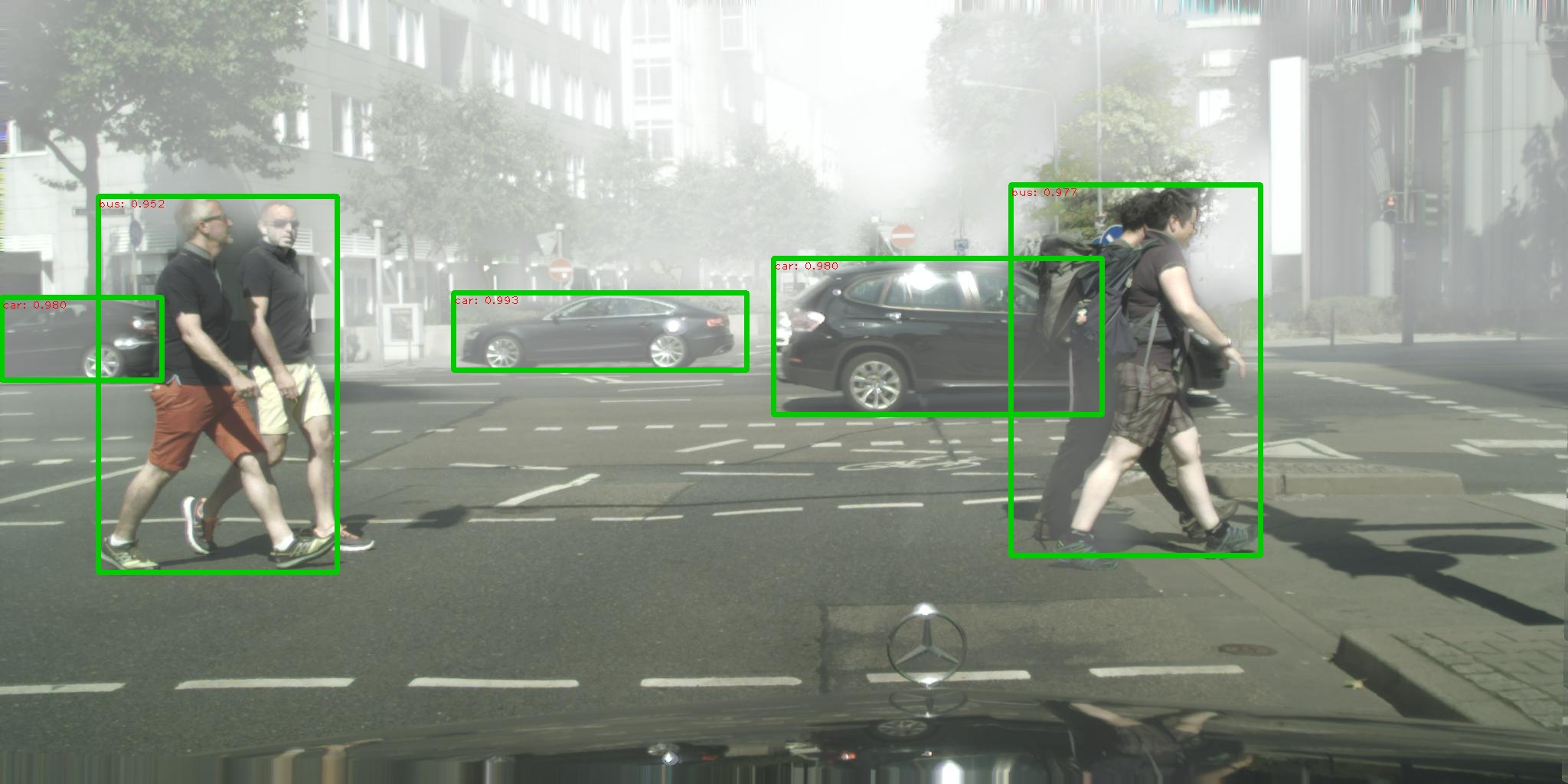}
\end{subfigure}
\begin{subfigure}{.32\linewidth}
  \centering
  \includegraphics[width=\linewidth]{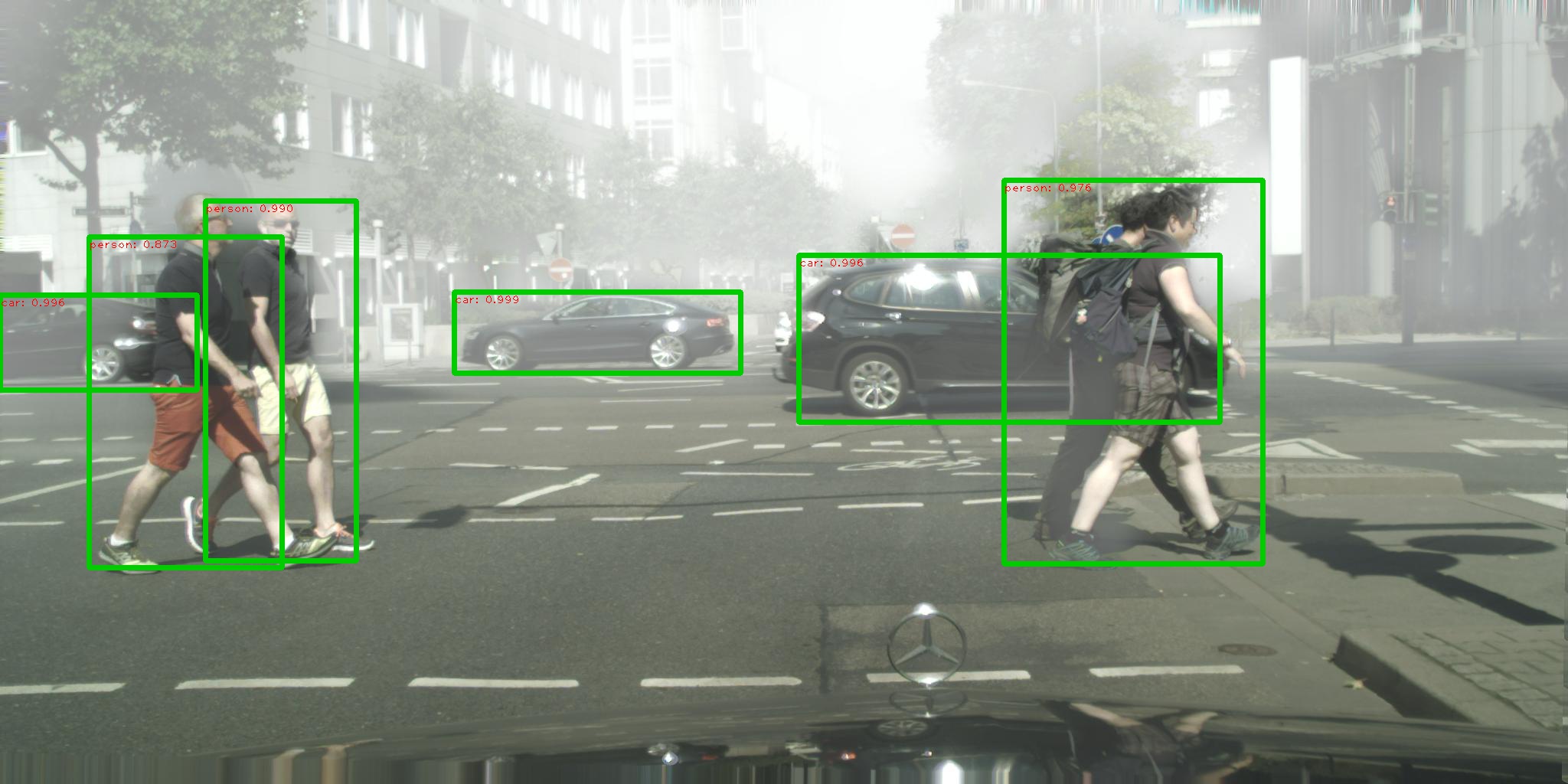}
\end{subfigure}
\\[\smallskipamount]

\begin{subfigure}{.32\linewidth}
  \centering
  \includegraphics[width=\linewidth]{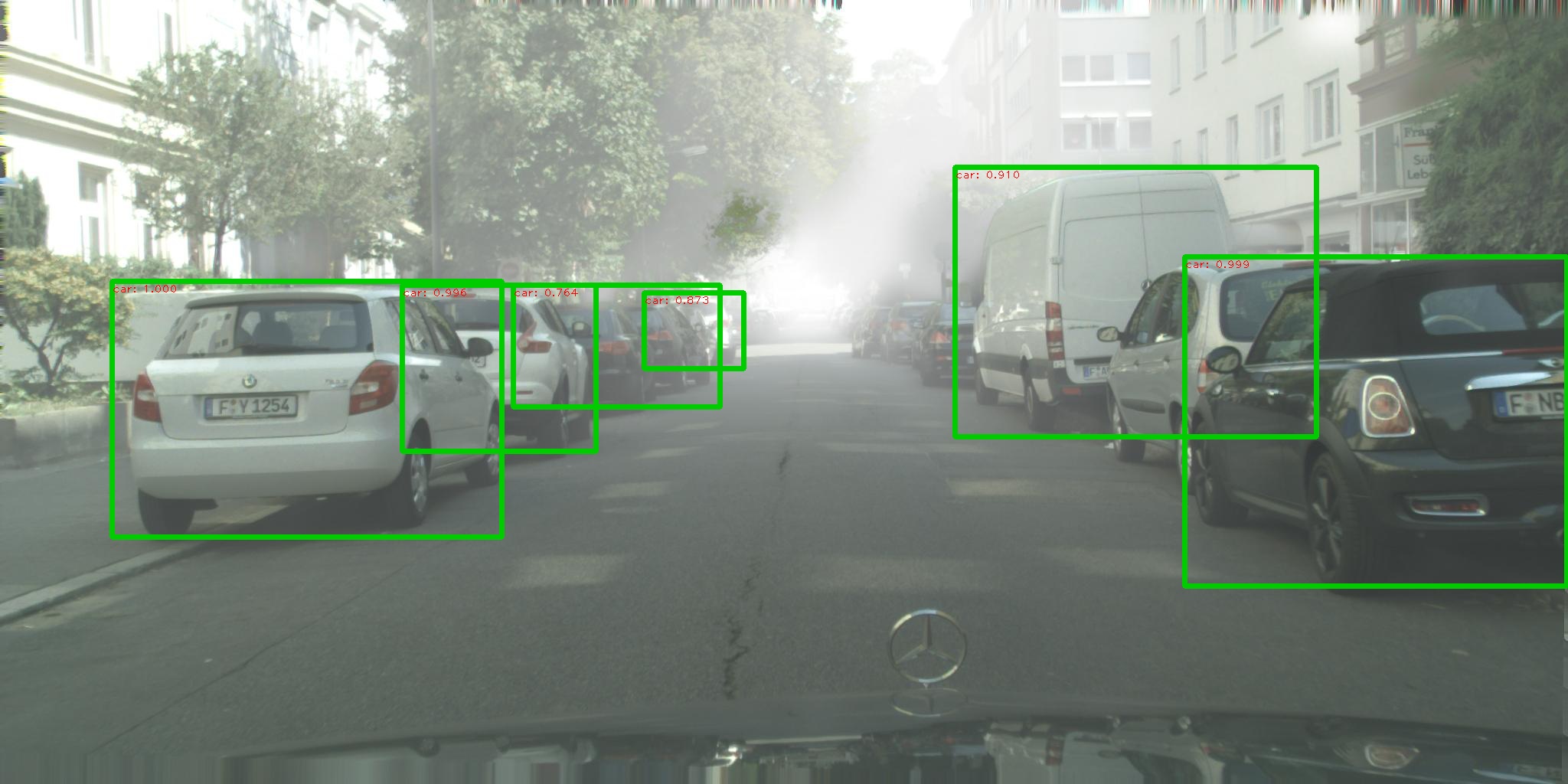}
\end{subfigure}
\begin{subfigure}{.32\linewidth}
  \centering
  \includegraphics[width=\linewidth]{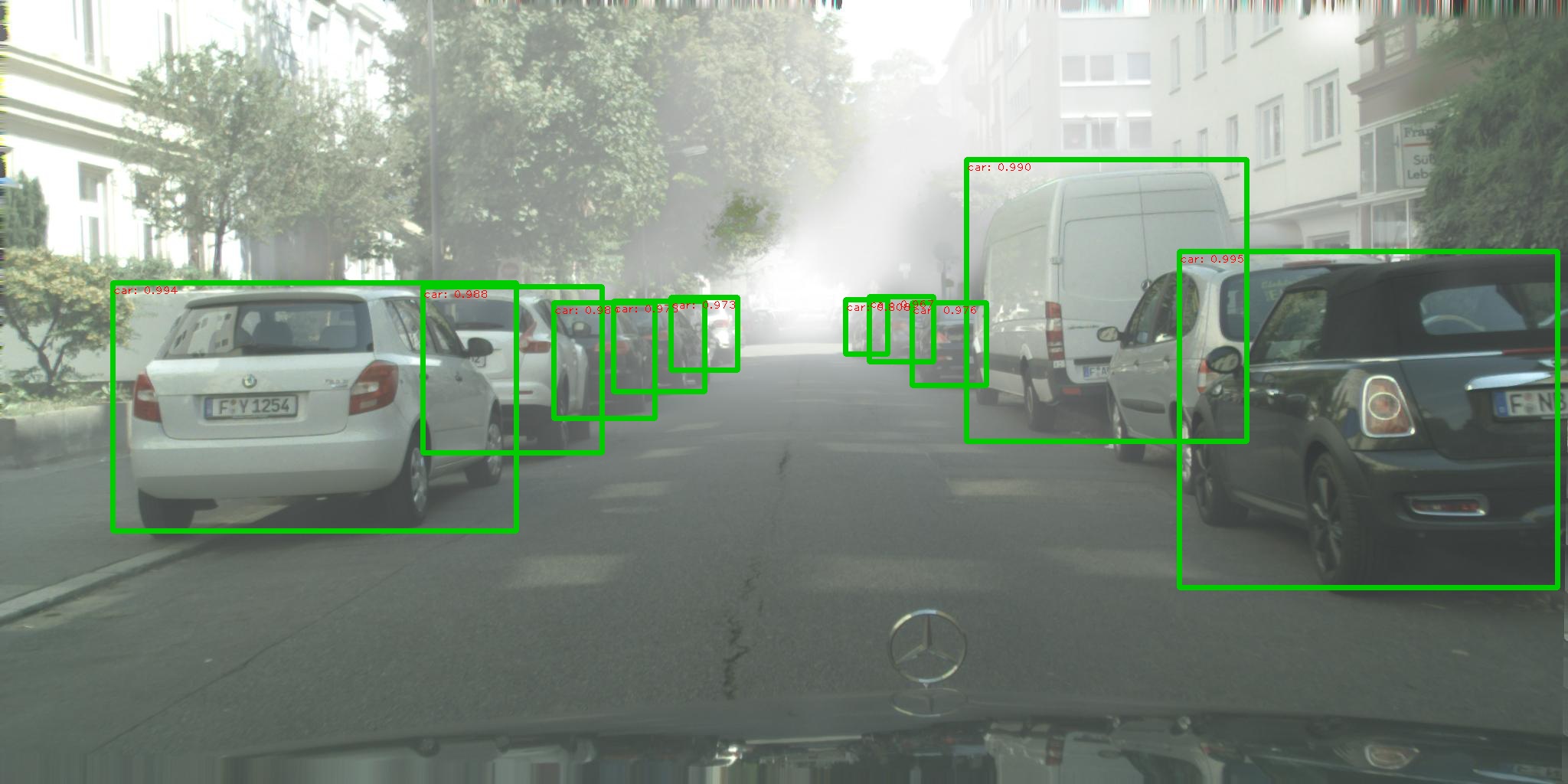}
\end{subfigure}
\begin{subfigure}{.32\linewidth}
  \centering
  \includegraphics[width=\linewidth]{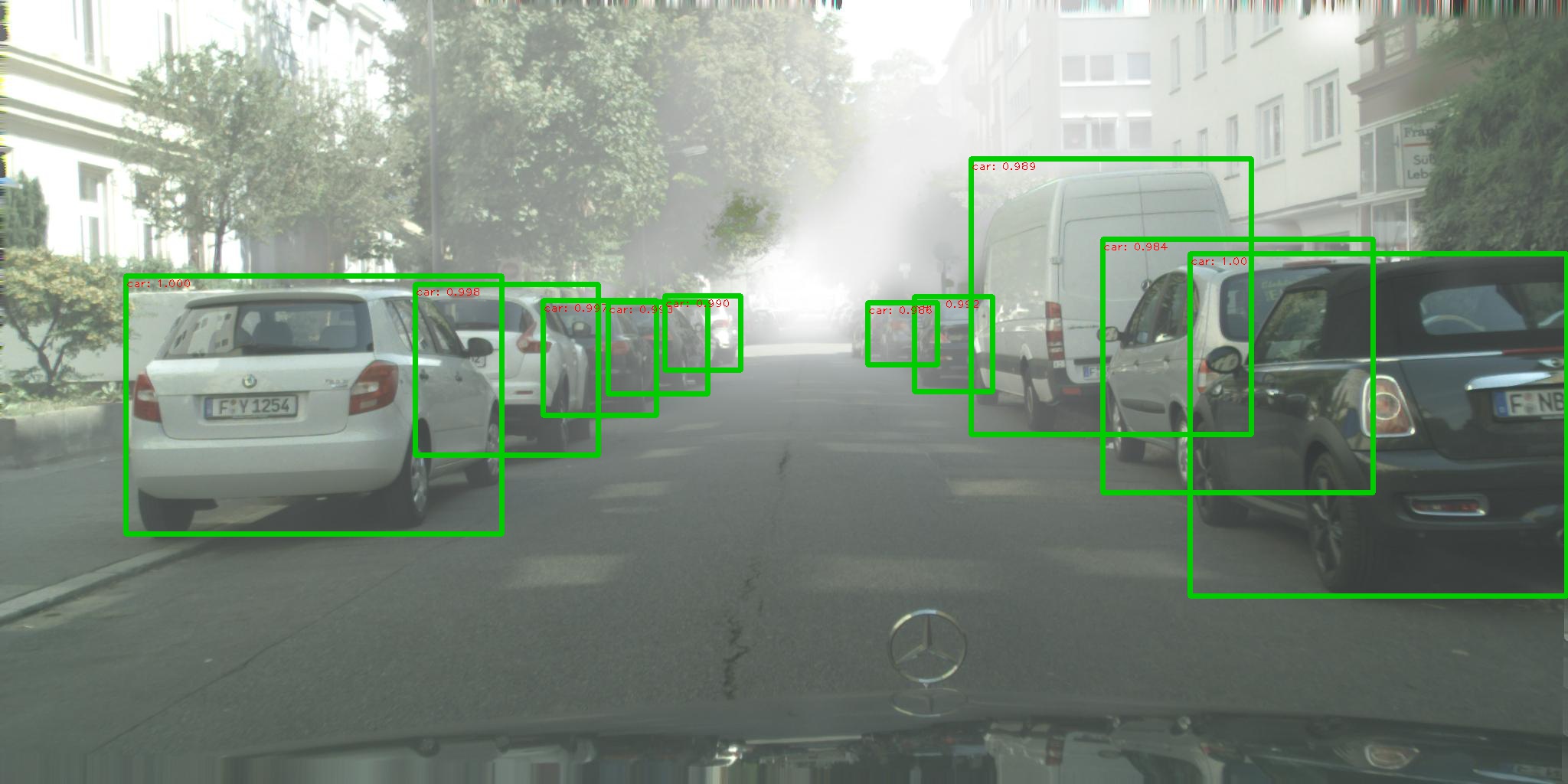}
\end{subfigure}
\\[\smallskipamount]

\begin{subfigure}{.32\linewidth}
  \centering
  \includegraphics[width=\linewidth]{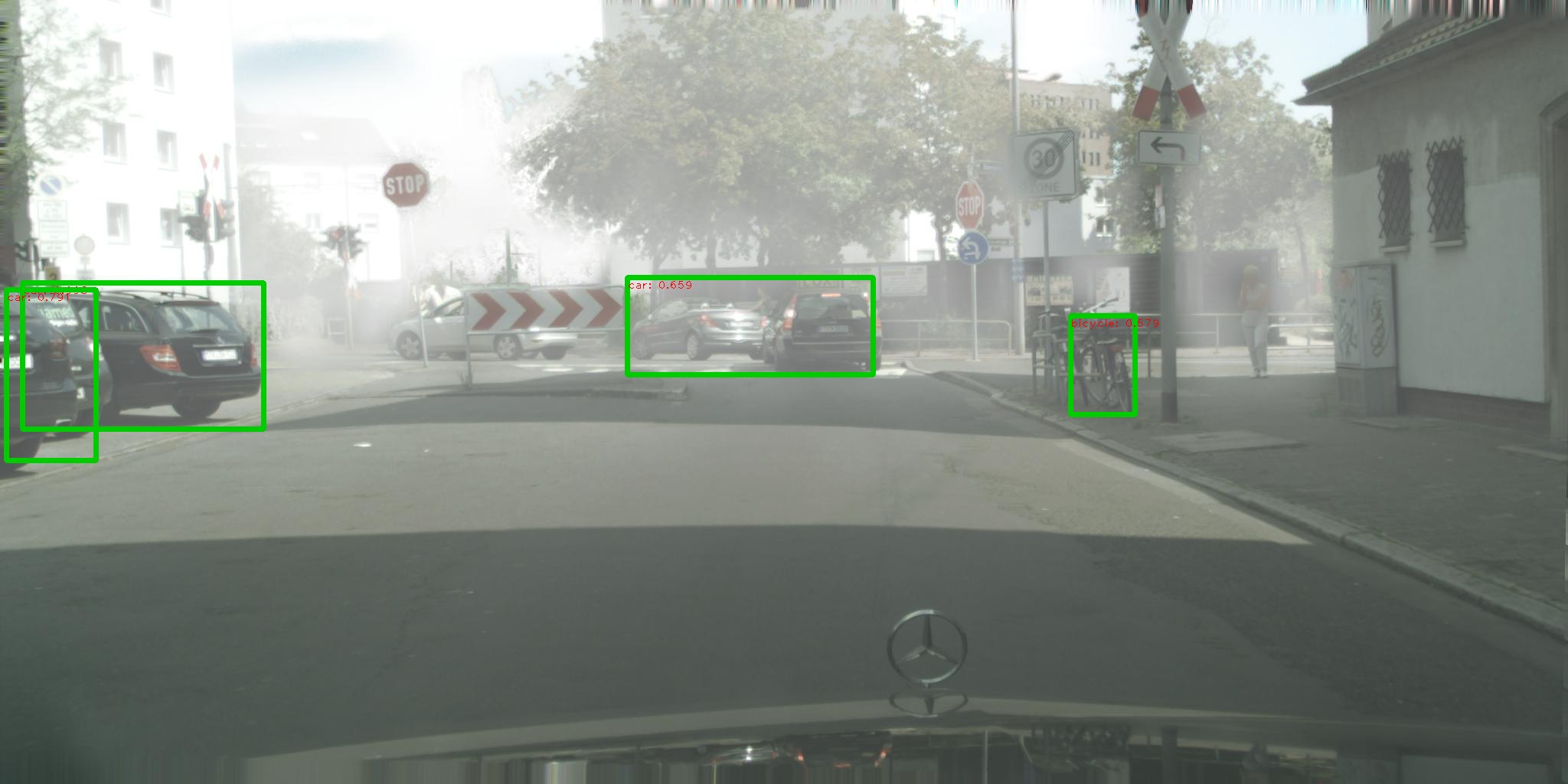}
\end{subfigure}
\begin{subfigure}{.32\linewidth}
  \centering
  \includegraphics[width=\linewidth]{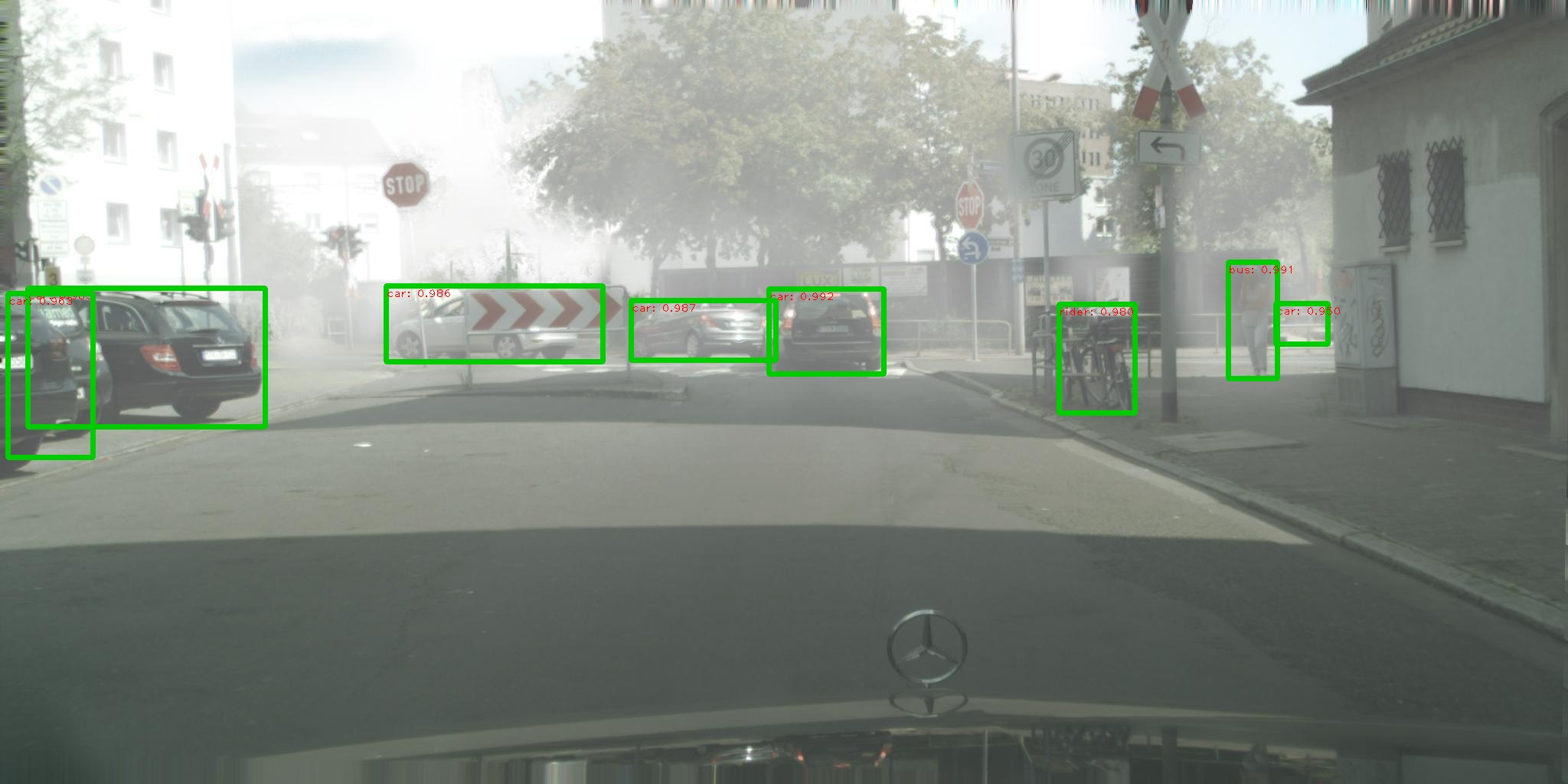}
\end{subfigure}
\begin{subfigure}{.32\linewidth}
  \centering
  \includegraphics[width=\linewidth]{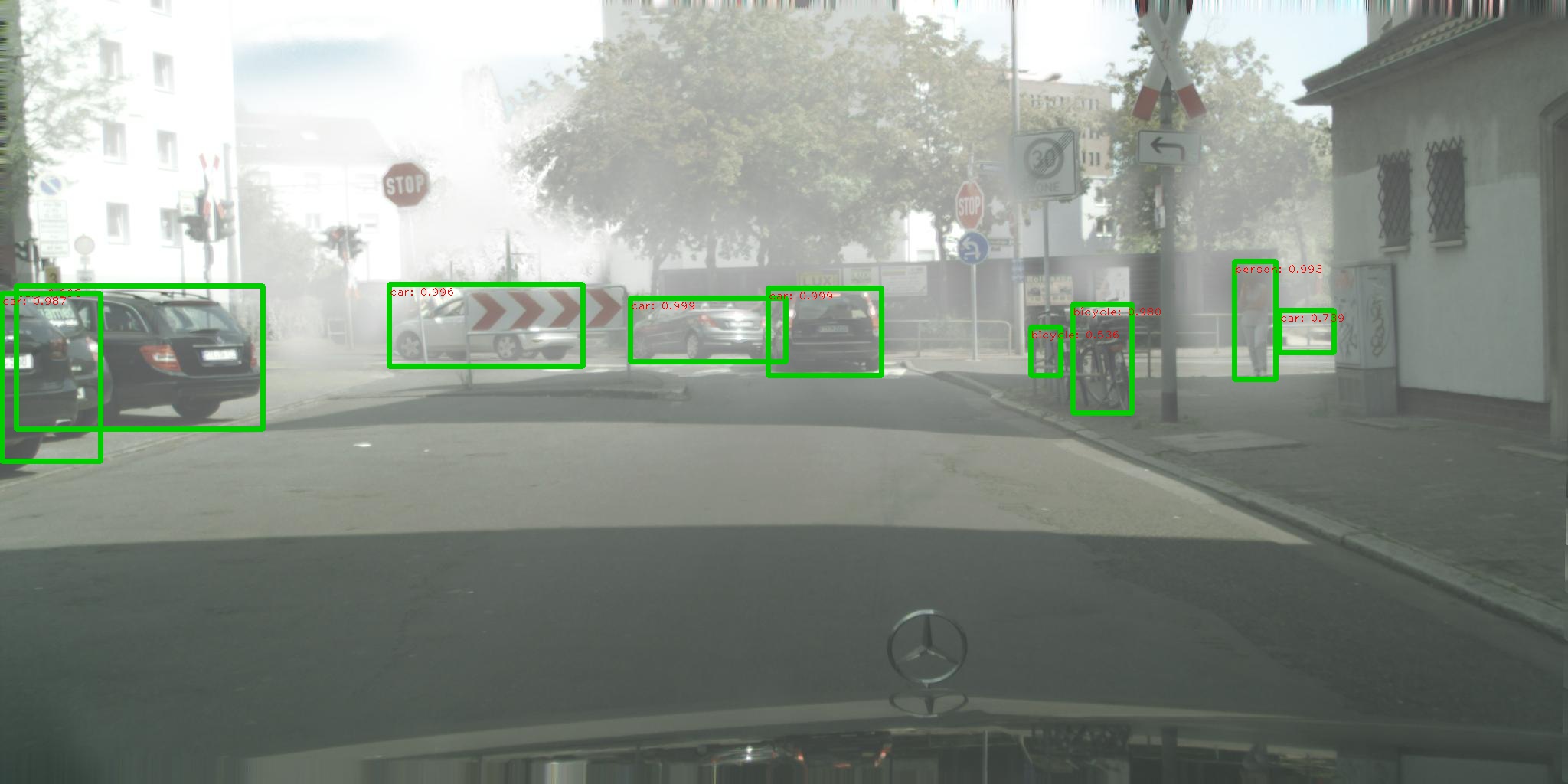}
\end{subfigure}
\\[\smallskipamount]

\begin{subfigure}{.32\linewidth}
  \centering
  \includegraphics[width=\linewidth]{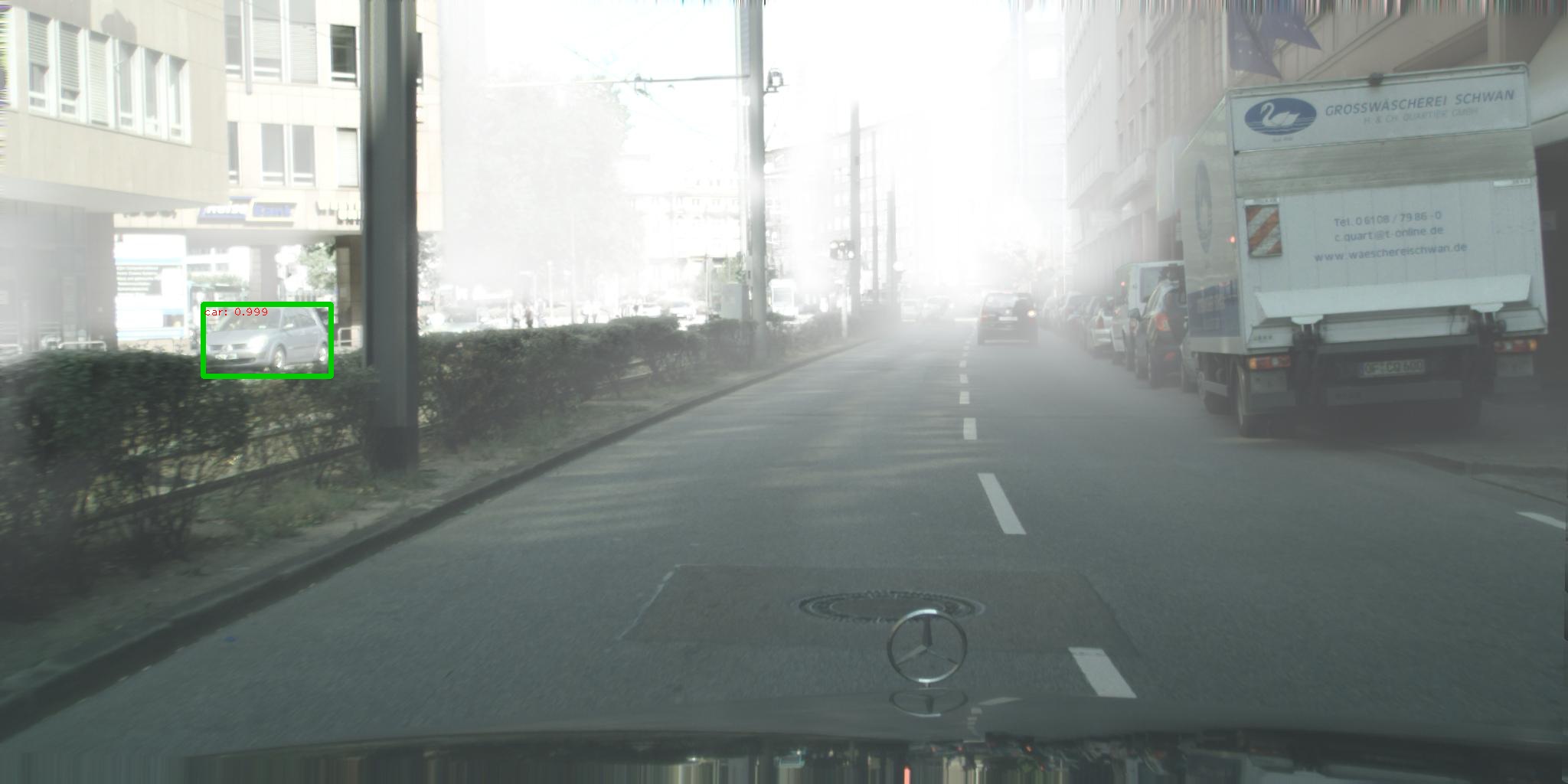}
  \caption{Source Only}
\end{subfigure}
\begin{subfigure}{.32\linewidth}
  \centering
  \includegraphics[width=\linewidth]{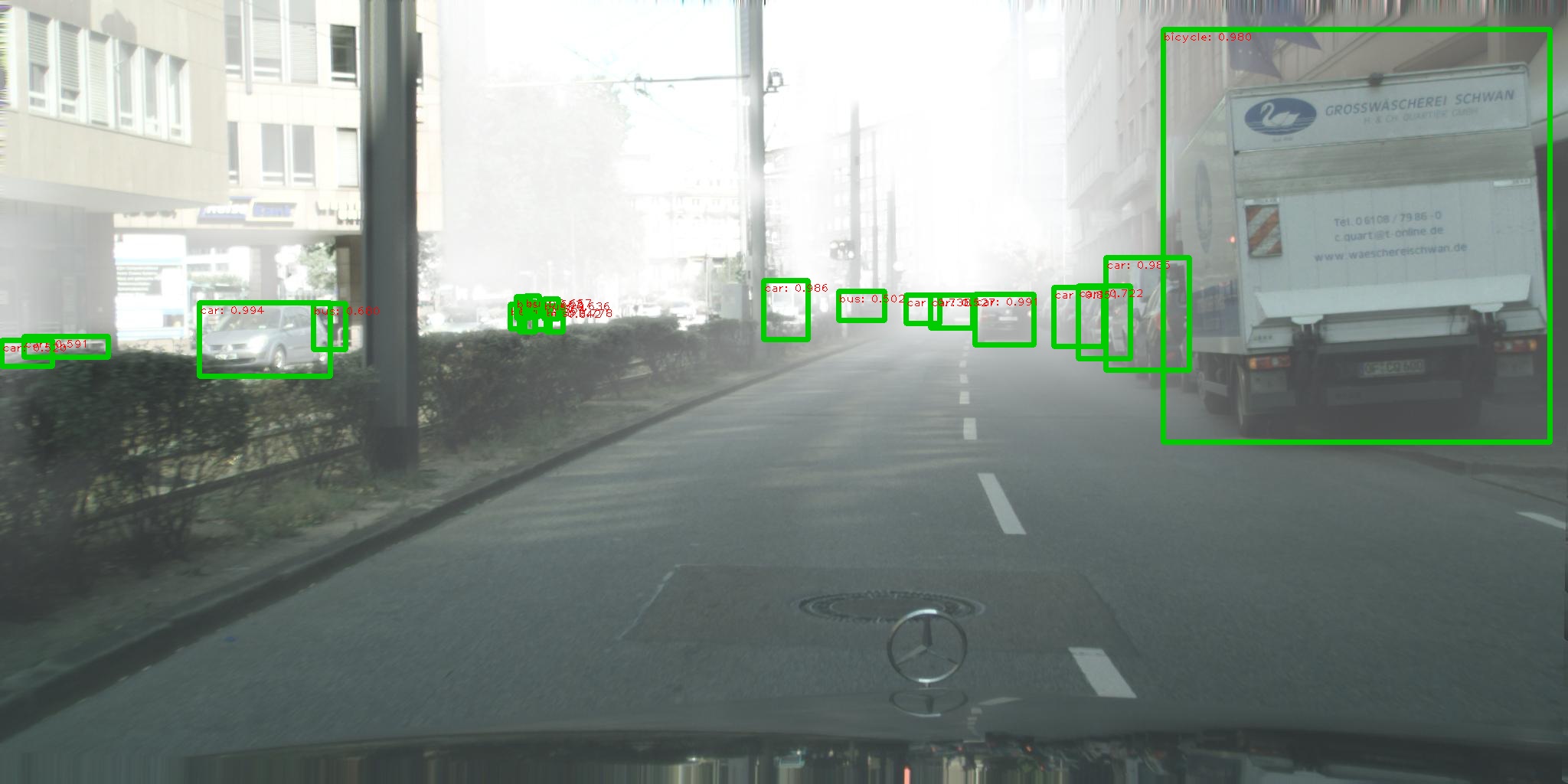}
  \caption{HTCN~\cite{htcn}}
\end{subfigure}
\begin{subfigure}{.32\linewidth}
  \centering
  \includegraphics[width=\linewidth]{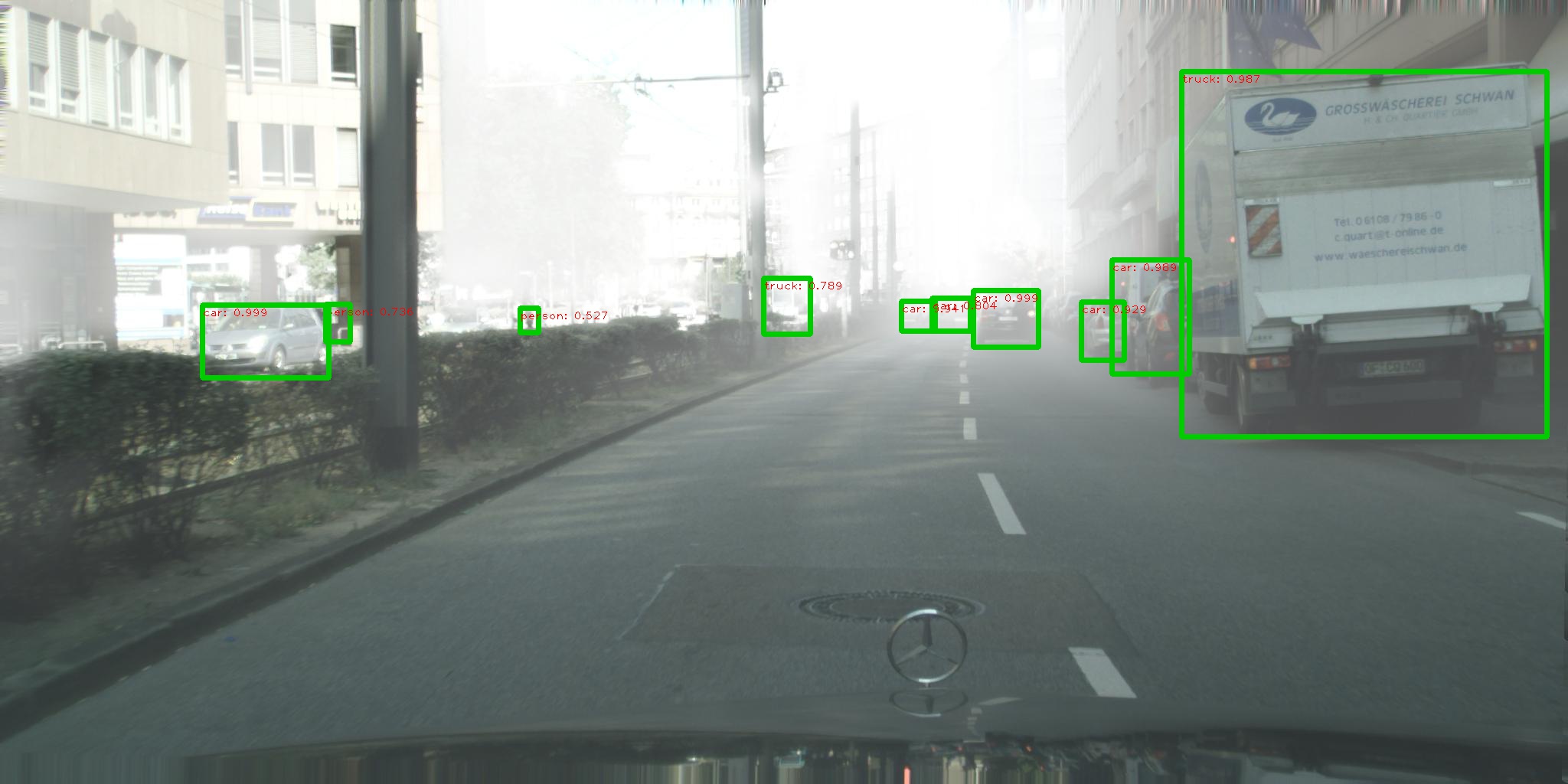}
  \caption{Our TIA}
\end{subfigure}
\\[\smallskipamount]

\caption{
Illustration of the detection results on the Cityscapes $\rightarrow$ Foggy Cityscapes benchmark.
Our TIA identifies more objects and delivers more accurate bounding boxes.
}
\label{vis2}
\end{figure*}

We provide some detection results of vanilla detector (\ie \textit{Source Only}~\cite{fasterrcnn}),
state-of-the-art adaptive detectors (\eg \textit{HTCN}~\cite{htcn} and \textit{UMT}~\cite{umt}),
and our framework TIA.
\cref{vis1} illustrates the comparison of detections on the PASCAL VOC~\cite{pascal} $\rightarrow$ Clipart~\cite{dt&pl} benchmark.
It is observed that our proposed TIA outperforms both \textit{Source Only} and \textit{UMT}~\cite{umt},
and produces more accurate detection results, \ie,
more foreground objects are identified (Row 1\&2), and higher quality bounding boxes are provided along with accurate categorization (Row 3-5).
Qualitative results on the Cityscapes~\cite{cityscape} $\rightarrow$ Foggy Cityscapes~\cite{foggycityscape} benchmark represented by~\cref{vis2}
also demonstrates the superiority of our TIA.
For example, in the first row, for the two cars on the left, the bounding box given by \textit{HTCN} is relatively off-target, 
while ours method present more compact boundaries, compared to \textit{Source Only}'s.

\section{Limitations}

The discrepancy between source and target domains in the label space, \ie, label shift, 
substantially affects the design philosophy and severely limits the performance of existing domain adaptive detectors.
In this subsection, we will provide in-depth analysis of how label shift limits our TIA for each dataset benchmark.

The benchmarks used in \textit{Normal-to-Foggy} (Cityscapes~\cite{cityscape} $\rightarrow$ Foggy Cityscapes~\cite{foggycityscape}) 
and \textit{Real-to-Artistic} (PASCAL VOC~\cite{pascal} $\rightarrow$ Clipart~\cite{dt&pl}) are essentially appropriate and 
they allow a good evaluation of the performance of various domain adaptive detectors.
Specifically, the former case is ideal, since it
shares an \textbf{identical label space} between the source and target domains, 
while the latter one has its label shift diluted due to the scale of the source domain.
In this context, it is observed that, our framework exceeds the upper bound indicated by \textit{Target Only} on the former benchmark 
and easily achieves state-of-the-art performance on the latter benchmark.

It is quite different in the \textit{Cross-Camera} scenario.
We find that the label shift of the benchmark (KITTI~\cite{kitti} $\leftrightarrow$ Cityscapes) employed in this scenario 
is dominated by the imbalance in the foreground-background ratio, namely
the inconsistency in the average number of objects between the source and target domain data.
In fact, the average numbers of instances of Cityscapes and KITTI are 9.1 and 3.8, respectively.
This directly leads to two serious problems.
On the one hand, we observe that the \textit{Source Only} model undergoes severe overfitting issue during training, 
which means that we underestimate the lower bound of the benchmark; 
on the other hand, it imposes higher demands on the cross-domain performance of RPN, 
and this straightforwardly undermines the effectiveness of the existing mainstream approaches that focus on feature alignment for it.

In summary, two arguments are made.
First, existing methods are highly inefficient in coping with label shift. 
In light of ~\cite{relax}, although the execution of domain alignment alone 
reduces the divergence between domains (the second term in Theorem 1), it
leads to arbitrary increases in $\lambda ^{\ast }$ (the third term in Theorem 1),
hence eventually,
the target errors of detectors cannot be well-guaranteed.
For this reason, taking into account the detectors' empirical predictions on the target domain,
or namely, the behavior of label predictors, is gradually emerging as a necessity.
Moreover, compared to classification tasks, the label shift in object detection task is considerably complicated.
It is no longer limited to the differences in category proportions, 
but is more widely distributed in spatial differences in scale, position, etc. of bounding boxes.
These two facts drive the proposal of TIA on a different aspect.

Second, in view of the fact that the label shift cannot be well estimated nor truly eliminated, 
we argue that there is a gap between the true upper bound and the present upper bound specified by \textit{Target Only}, 
according to ~\cite{label}. Under such circumstances, 
the close performance of the domain adaptive detectors in the \textit{Cross-Camera} benchmark can be reasonably explained.

\end{document}